\documentclass[letterpaper,twocolumn,10pt]{article}
\usepackage{usenix-2020-09}

\usepackage[utf8]{inputenc} 
\usepackage[T1]{fontenc}    
\usepackage{xurl}           
\usepackage[normalem]{ulem} 

\usepackage{booktabs}       
\usepackage{amsfonts}       
\usepackage{nicefrac}       
\usepackage{microtype}      
\usepackage{xcolor}         

\usepackage{epsfig}

\usepackage{graphicx}
\usepackage{array}
\usepackage{comment}

\usepackage{amsmath}
\usepackage{amssymb}
\usepackage{pifont}

\usepackage[accsupp]{axessibility}  

\usepackage[utf8]{inputenc}
\usepackage[utf8]{luainputenc}

\usepackage{ifthen}

\usepackage{nicefrac}

\usepackage{authblk}

\definecolor{tabblue}{HTML}{1F77B4}
\definecolor{taborange}{HTML}{FF7F0E}
\definecolor{tabgreen}{HTML}{2CA02C}

\newif\ifcomments
\commentsfalse
\ifcomments
\usepackage[textsize=small,color=lightgray]{todonotes}
\else
\usepackage[disable]{todonotes}
\fi

\usepackage{textcomp}

\usepackage{pgfplots}
\pgfplotsset{compat=1.13}

\usepackage{enumitem}

\usepackage{etoolbox}
\makeatletter
\patchcmd{\ttlh@hang}{\parindent\z@}{\parindent\z@\leavevmode}{}{}
\patchcmd{\ttlh@hang}{\noindent}{}{}{}
\makeatother

\usepackage{balance}

\usepackage{datetime}

\usepackage{listings}

\usepackage[ruled,vlined,linesnumbered]{algorithm2e}

\SetCommentSty{mycommfont}

\usepackage[capitalize]{cleveref}
\crefname{section}{Sec.}{Secs.}
\Crefname{section}{Section}{Sections}
\Crefname{table}{Table}{Tables}
\crefname{table}{Tab.}{Tabs.}
\Crefname{listing}{Listing}{Listings}
\crefname{listing}{Lst.}{Lst.}
\AtBeginEnvironment{subappendices}{\crefalias{section}{appendix}}

\usepackage{xfrac}

\usepackage{rotating}
\usepackage{xparse}

\usepackage{outlines}

\usepackage{xspace}
\usepackage{framed}

\usepackage[page,toc]{appendix}
\usepackage{titletoc}

\usepackage{setspace}

\usepackage{adjustbox}

\usepackage{mdframed}

\newmdenv[
  backgroundcolor=yellow!20,
  linecolor=yellow!50!black,
  skipabove=10pt,
  skipbelow=10pt,
  leftmargin=0pt,
  rightmargin=0pt,
  innerleftmargin=10pt,
  innerrightmargin=10pt,
  innertopmargin=10pt,
  innerbottommargin=10pt
]{highlightbox}

\newboolean{draft}
\setboolean{draft}{false}

\ifcomments
\paperwidth=\dimexpr \paperwidth + 7cm\relax
\oddsidemargin=\dimexpr\oddsidemargin + 4cm\relax
\evensidemargin=\dimexpr\evensidemargin + 4cm\relax
\marginparwidth=\dimexpr \marginparwidth + 4cm\relax
\fi

\newcommand{\eyal}[1]{\todo[color=gray!40]{\textbf{Eyal}: #1}\ignorespaces}

\newcommand{\myparagraph}[1]{\noindent{\bfseries #1.}}

\NewDocumentCommand{\rot}{O{45} O{1em} m}{\makebox[#2][l]{\rotatebox{#1}{#3}}}

\usepackage{comment}

\usepackage{color}
\usepackage{color, soul}

\ifdraft
 \usepackage{type1cm}
 \usepackage{eso-pic}
 \makeatletter
 \AddToShipoutPicture{%
             \setlength{\@tempdimb}{.5\paperwidth}%
             \setlength{\@tempdimc}{.53\paperheight}%
             \setlength{\unitlength}{1pt}%
             \put(\strip@pt\@tempdimb,\strip@pt\@tempdimc){%
         \makebox(0,700){\textcolor[gray]{0.50}%
         {\fontsize{15pt}{15pt}\selectfont{Draft of \today{}, {\currenttime} -- please do not distribute.}}}%
         \makebox(0,0){\rotatebox{45}{\textcolor[gray]{0.92}%
         {\fontsize{4cm}{4cm}\selectfont{DRAFT}}}}%
             }%
 }
\fi

\usepackage[capitalize]{cleveref}
\crefformat{section}{#2\S#1#3}

\DeclareMathAlphabet{\mathcal}{OMS}{cmsy}{m}{n}

\newcommand{\one}{({\em i}\/)}
\newcommand{\two}{({\em ii}\/)}
\newcommand{\three}{({\em iii}\/)}

\definecolor{custompurple}{rgb}{0.58, 0.13, 0.57}
\definecolor{customgreen}{rgb}{0.11, 0.69, 0.0}

\newcommand{\system}{RedunCut}

\newcommand{\DMSS}{DMSS}

\usepackage[labelfont=bf]{caption}
\usepackage[aboveskip=2pt,belowskip=1pt]{subcaption}

\newlength{\oldtextfloatsep}
\newlength{\oldintextsep}

\usepackage{multirow}

\usepackage{float}
\usepackage{afterpage}

\usepackage{makecell}

\newenvironment{tightitemize}{%
\begin{list}{$\bullet$}{%
\setlength{\itemsep}{1.5pt}%
\setlength{\topsep}{2pt}%
\setlength{\parskip}{0pt}%
\setlength{\parsep}{0pt}%

\setlength{\labelwidth}{0pt}%
\setlength{\leftmargin}{4pt}%
\setlength{\labelsep}{0pt}%
\setlength{\listparindent}{0pt}%
}}%
{\end{list}}

\newenvironment{tightenumerate}{%
\begin{list}{\labelenumi}{%
\usecounter{enumi}%
\setlength{\itemsep}{1.5pt}%
\setlength{\topsep}{2pt}%
\setlength{\parskip}{0pt}%
\setlength{\parsep}{0pt}%

\setlength{\labelwidth}{0pt}%
\setlength{\leftmargin}{4pt}%
\setlength{\labelsep}{0pt}%
\setlength{\listparindent}{0pt}%
}}%
{\end{list}}

\usepackage{tikz}
\usepackage{amsmath}

\usepackage{filecontents}

\usepackage{titlesec}
\titlespacing*{\section}{0pt}{1ex}{1ex} 

\begin{document}

\date{}

\title{\Large \bf \system{}: Measurement-Driven Sampling and Accuracy Performance Modeling for Low-Cost Live Video Analytics}


\author[1]{Gur-Eyal Sela}
\author[1]{Kumar Krishna Agrawal}
\author[2]{Bharathan Balaji$^{*}$\thanks{Work unrelated to Amazon}}
\author[1]{Joseph Gonzalez}
\author[1]{Ion Stoica}
\affil[1]{UC Berkeley}
\affil[2]{Amazon}


\maketitle

\begin{abstract}
Live video analytics (LVA) runs continuously across massive camera fleets, but
inference cost with modern vision models remains high. To address this, dynamic
model size selection (DMSS) is an attractive approach: it is content-aware but
treats models as black boxes, and could potentially reduce cost by up to
10$\times$ without model retraining or modification. Without ground truth
labels at runtime, we observe that DMSS methods use two stages per segment: (i)
sampling a few models to calculate prediction statistics (e.g., confidences),
then (ii) selection of the model size from those statistics. Prior systems fail
to generalize to diverse workloads, particularly to mobile videos and lower
accuracy targets. We identify that the failure modes stem from inefficient
sampling whose cost exceeds its benefit, and inaccurate per-segment accuracy
prediction.

In this work, we present RedunCut, a new DMSS system that addresses both: It
uses a measurement-driven planner that estimates the cost-benefit tradeoff of
sampling, and a lightweight, data-driven performance model to improve accuracy
prediction. Across road-vehicle, drone, and surveillance videos and multiple
model families and tasks, RedunCut reduces compute cost by 14-62\% at fixed
accuracy and remains robust to limited historical data and to drift. 
\end{abstract}

\section{Introduction}
\label{s:introduction}

Live video analytics (LVA) performs continuous inference on camera streams to
detect events and trigger real-time actions. It now underpins public safety,
traffic, retail, industrial, and consumer applications
\cite{datondji2016survey,cangialosi2022privid,securityinformed_cameras2024,aws_smartstore_analytics,voxel51_retail_cv,ekya},
with deployments spanning millions of concurrent video streams across
edge-cloud
infrastructure~\cite{ekya,recl,ananthanarayanan2017real,clarion_london_cctv_2022,solink_video_bigdata,nyt_chicago_surveillance_2018,qz_beijing_surveillance_2015}.
Since state-of-the-art vision model inference is expensive per
frame~\cite{lin2017feature,efficientdet,cai2015learning,ge2021yolox,carion2020end,recl,howard2017mobilenets,efficientvit}
and total cost grows roughly linearly with video volume (number of videos,
frame rates, duration)~\cite{jain2019scaling}, the cost of processing a single
30 fps video stream can exceed \$1,000/month~\cite{recl}. At scale, the
accuracy-cost tradeoff dominates system design.

Beyond growth in scale and cost, LVA applications have broadened along two
axes. \underline{First}, sources are increasingly mobile: improvements in
cellular and Wi-Fi
networks~\cite{ni2023cellfusion,xu2022tutti,gsma2022drones5g} enable inference
on video feeds from road vehicles and
UAVs~\cite{wong2024madeye,wang2022enabling,nguyen2022state,nsc_adas_stats},
extending existing uses (e.g., license-plate queries for AMBER alerts, on-road
hazard detection~\cite{yi2017lavea,zhang2019edge,elsahly2022systematic}) and
enabling new ones (e.g., search-and-rescue, wildfire spotting, crop monitoring
\cite{lyu2023unmanned,pesonen2025boreal,EPA_PPDC_EmergingAgTech_2021}).
\underline{Second}, accuracy targets span a wide range: safety-critical tasks
demand high accuracy (e.g., weapon/intruder detection, drone-assisted
search-and-rescue
\cite{debnath2021comprehensive,usf_darts_2024,ciccone2025real}), whereas
cost-sensitive analytics tolerate approximate answers and aggregates (e.g.,
foot-traffic/occupancy estimation~\cite{aws_smartstore_analytics},
signage/lane-marking defects and pothole detection from
dashcams~\cite{safyari2024review}). Consequently, an LVA system must serve both
ends, delivering high accuracy as well as inexpensive, approximate analytics
when requested.

\textbf{Goals.} Our goal is to create an LVA system that minimizes compute cost
while meeting the application's specified accuracy target. This should be done
while serving applications running on a range of video workload types (mobile
and stationary) as well as a range of accuracy targets (low to high).

\textbf{Methods to improve LVA.} Existing methods to improve LVA can be
classified as \one{} \textit{content-agnostic} and \two{}
\textit{content-aware}. \underline{Content-agnostic} methods ignore video
content and optimize cost and accuracy via dynamic batching, placement, and
accelerator tuning~\cite{clipper,nexus,alpa,infaas}. They use the same model
size throughout the video (\textit{static model size selection}) and do not
adapt resource allocation to the video content, missing cost-reduction
opportunities. To fill this gap, recent \underline{content-aware} works adapt
resource allocation across different parts of the video based on content (e.g.,
dynamic model training~\cite{ekya,recl}, early-exit
mechanisms~\cite{E3,apparate}, and mixture of
experts~\cite{moe_1,li2023accelerating}). While these works achieve greater
cost reduction, many modify base model weights or architectures and require
significant machine-learning expertise to port to new tasks, workloads, and
model types.

\textbf{Opportunity: Dynamic model size selection (DMSS).} In contrast,
\textit{dynamic model size selection} is a different content-aware method that
treats the models as black boxes (no modification), and allows the use of
off-the-shelf pretrained model zoos, which simplifies development, reduces
engineering effort, and speeds deployment while still meeting the application's
accuracy target. In \DMSS{}, the video is divided into segments (e.g., 2
seconds~\cite{chameleon}), and a different model size is used to process each
video segment. This approach exploits the fact that the accuracy of different
model sizes changes across video segments. A smaller, inexpensive model is used
on video segments where its accuracy is high, while the larger model is
reserved for the rest of the video segments. For example, with an accuracy
target of 33.9 MOTA, EDet-D3 (a ``small'' model with average accuracy of 27.08)
still exceeds the target on 31.7\% of the video segments, and can be selected
in those video segments. Indeed, we find that using \DMSS{} can achieve up to
10$\times$ lower cost than static model size selection while preserving the
accuracy target (\cref{s:background-motivation}).

\begin{figure}[t!]
    \centering
    \captionsetup[subfigure]{width=0.99\textwidth}
    \includegraphics[width=0.48\textwidth]{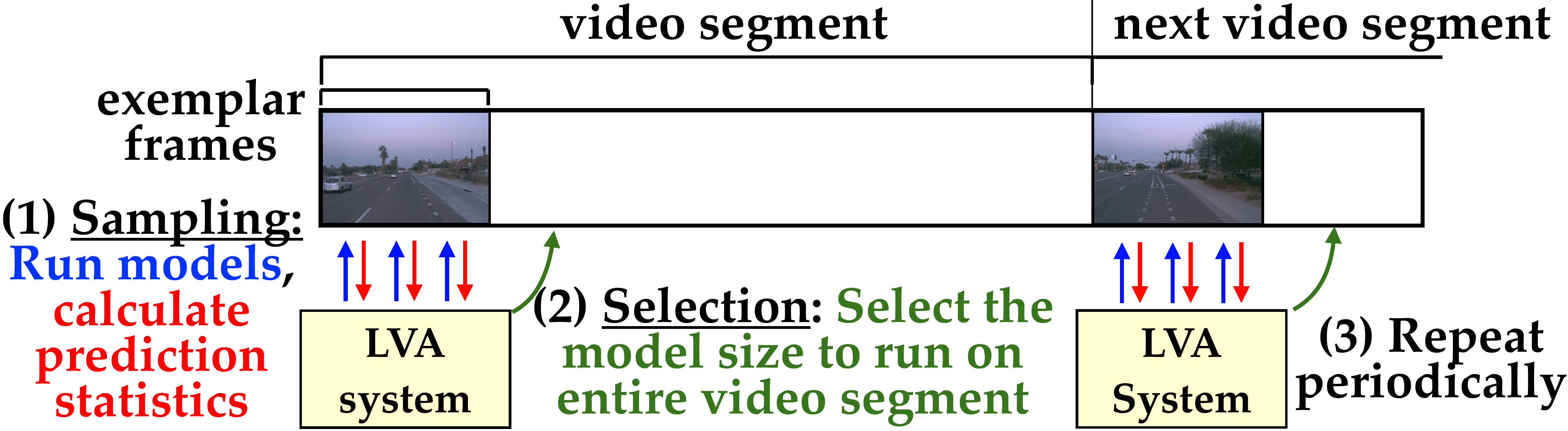}
    \caption{\small \textbf{Periodic sampling and selection.}
    }
    \label{f:runtime-profiling-setup}
    \vspace{-1em}
\end{figure}

\textbf{The use of prediction statistics in \DMSS{}.} Selecting the best model
size in each video segment requires determining the accuracy of all candidate
model sizes on each video segment at runtime. The optimal solution, which runs
all models on all frames and compares the predictions to ground truth labels
(e.g., human annotations), cannot be used: First, it costs more than a static
selection method that runs the largest, most-accurate model alone. Second, the
ground truth labels are infeasible or too expensive to access at runtime.
Instead, current systems combine two alternatives
(\Cref{f:runtime-profiling-setup}). First, they evaluate only a few
\textit{exemplar frames} (e.g., the first few frames) and extrapolate accuracy
to the rest of the segment. Second, they replace using ground-truth labels with
properties of model predictions that correlate with the accuracy (e.g.,
confidence, entropy, disagreement score, sum of objectness
scores~\cite{cascade,tabi,noscope,ekya,recl,videostorm,dds,EAAR}). We
collectively refer to these properties as \textit{prediction statistics}.
Together, the processing of each video segment at runtime can be broken into
two stages: \one{} \textit{sampling}: one or more of the model sizes are
sampled by running on exemplar frames to get their predictions, and using these
predictions to calculate the prediction statistics. \two{} \textit{selection}:
These prediction statistics are used to select the model size that optimizes
cost and accuracy on the entire video segment. This process is repeated in each
video segment to adapt to changing video content.

\textbf{Challenges when using prediction statistics: sampling and selection
(\cref{s:background-motivation}).}

\noindent 1. \textit{What is the optimal subset of model sizes to
\textbf{sample} on the exemplar frames to calculate prediction statistics?}
Sampling too many or too few model sizes may both result in prohibitively high
cost. The optimal number of models to sample also widely varies across video
segments.

\noindent 2. \textit{Given only a subset of prediction statistics, which model
size to \textbf{select} to run on the entire video segment?} The decision must
be made from prediction statistics that may only weakly correlate with the
accuracy.

\textbf{Existing works} use manually-crafted rules for both steps: sample all
model sizes once then fix a size for the rest of the
video~\cite{remix,videostorm}; cascades that sample from small to large with
early exits~\cite{noscope,tahoma,tabi,cascade,dds}; or more elaborate rule
sets~\cite{chameleon}. These manually-crafted rules are tuned for surveillance
workloads and high-accuracy targets. However, in mobile-camera workloads (e.g.,
road vehicles, drones) and at lower accuracy targets, we see up to 75\% higher
cost than the static selection baseline. We isolate two factors that drive this
gap (\cref{s:solution-approach}):

\textbf{A. Inefficient sampling.} Sampling each additional model has two
opposing effects: \one{} Cost: Sampling increases cost, and \two{} Benefit:
Sampling may also reveal a prediction statistic that can be used to select a
better model size, e.g., a smaller, less expensive model that preserves
accuracy. Existing works do not estimate the \textit{net effect} of sampling
and often sample when the cost exceeds the expected benefit, resulting in a net
cost increase.

\textbf{B. Suboptimal selection.} The relationship between prediction
statistics and accuracy is weak and varies across video segments and workload
types. Using thresholds~\cite{noscope,remix,videostorm} and heuristic
rules~\cite{chameleon} misses this complexity, resulting in suboptimal model
size selection even when sampling is adequate.

\textbf{Implications.} A better approach must \one{} estimate the net value of
each potential model to sample, and \two{} better capture the relationship
between the prediction statistics and the accuracy to improve accuracy
prediction, and thereby model size selection.

To implement these design principles and achieve the cost-saving and simplicity
advantages of dynamic model size selection across different accuracy targets
and workload types, \underline{in this paper} we propose \system{}, a new live
video analytics system.

\textbf{Solution.} \underline{System overview:} \system{} dynamically selects
the model size for different segments throughout the video. It is a
model-agnostic shim-layer that bolts on top of any prediction-serving system,
and has a \textbf{control plane} (containing the \textit{sampler} and the
\textit{selector}) and a \textbf{data plane} that runs the selected model size.
The sampler runs a sequential, measurement-driven planner on the exemplar
frames to decide which models to sample, producing prediction statistics. The
selector uses these prediction statistics and a lightweight, learned accuracy
performance model to predict the accuracy, which is used to select the
best model size. The data plane runs the selected model size on the entire
video segment.

\underline{Sampling: measurement-driven planner.} \system{} uses the key
insight that prediction statistics of different model sizes correlate with each
other (\cref{s:solution-approach}). Leveraging this insight, \system{} uses the
prediction statistics of models that were sampled to narrow the range of the
possible values of prediction statistics of model sizes that were not sampled
yet. This narrower range of values is used by \system{}'s measurement-driven
planner to better estimate the net effect of sampling models that were not
sampled yet.

\underline{Selection: Accuracy performance model.} Prediction statistics are
imperfect \textit{proxy metrics} for accuracy, and their relationship to
accuracy varies across workloads and model families
(\cref{ss:improving-selection}). We therefore treat accuracy prediction as a
\textit{performance modeling} problem: we use a learned, lightweight predictor
that maps available prediction statistics to each model's accuracy, analogous
to systems that use purpose-built performance models to predict target metrics
(e.g., latency/throughput) from noisy proxy metrics at
runtime~\cite{paragon,quasar,cilantro}. \system{} uses a lightweight kNN model
that is robust to missing prediction statistics. Its accuracy predictions are
used to select the model size in each video segment.

\textbf{Evaluation.} We evaluate \system{} on instance segmentation and
multi-object tracking across six combinations spanning three model families
(EfficientDet~\cite{efficientdet}, YOLO~\cite{ge2021yolox},
DETR~\cite{zong2023detrs}), and three video workload types (surveillance, road
vehicles, drones). Compared with state-of-the-art
baselines~\cite{recl,tabi,chameleon,noscope}, \system{} lowers GPU cost by
14-62\% for the same target accuracy. \system{} remains robust when the
historical workload data \one{} is reduced by 8$\times$, \two{} undergoes data
drift, or \three{} when ground-truth labels are machine-generated. In ablation
studies, we find that on average 61\% of the cost reduction is achieved by
using the new sampling approach, while the rest of the improvement comes from
better accuracy prediction during selection. We also find that \system{}'s
accuracy prediction has $3.5-10\times$ lower error versus rules used in
existing works.

In summary, our paper's contributions are the following:
\begin{tightenumerate}
    \item To our knowledge, this is the first study to investigate dynamic
    model size selection in mobile video sources. We show that estimating the
    cost-benefit of sampling and learning the weak and variable relationships
    of prediction statistics and accuracy is essential to improve performance
    in such videos.
    \item We develop a system (\system{}) that integrates these insights to
    generalize to a wide variety of video workload types and accuracy targets.
    \item We evaluate this video analytics system and find that it reduces
    cost by 14-62\% over existing methods for the same accuracy target across
    different video workload types, model architectures, and tasks.





\end{tightenumerate}

\section{Background and Motivation}
\label{s:background-motivation}

\begin{figure}[t!]
    \centering
    \captionsetup[subfigure]{width=0.99\textwidth}
    \includegraphics[width=0.48\textwidth]{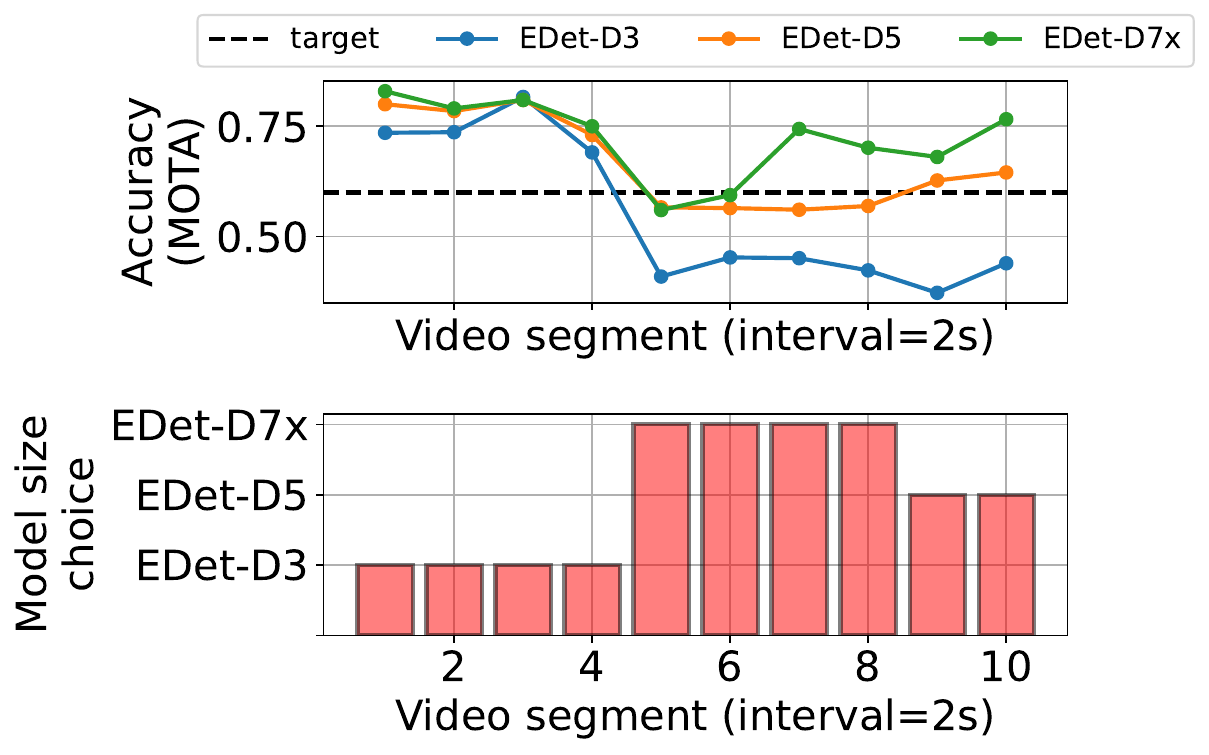}
    \caption{\small \textbf{The accuracy of various model sizes in a 20 second
            video (top), and the resulting model size selection (bottom).} Top:
            The accuracy (y-axis) of different model sizes (color) on 2-second
            video segments of a road vehicle video (x-axis). Bottom: The
            resulting model size selection (y-axis) for each video segment
            (x-axis).
        The models are taken from the EfficientDet
        (EDet)~\protect\cite{efficientdet} family.}
    \label{f:model-accs}
    \vspace{-1em}
\end{figure}

\textbf{System goals.} The LVA system aims to minimize the cost, i.e., the
total computational cost (e.g., GFLOPs) spent on processing the videos, while
meeting the application's accuracy target, i.e., the aggregate accuracy of the
predictions on the videos' frames.

\begin{figure*}[th!]
    \centering
    \captionsetup[subfigure]{width=0.99\textwidth}
    \includegraphics[width=1.0\textwidth]{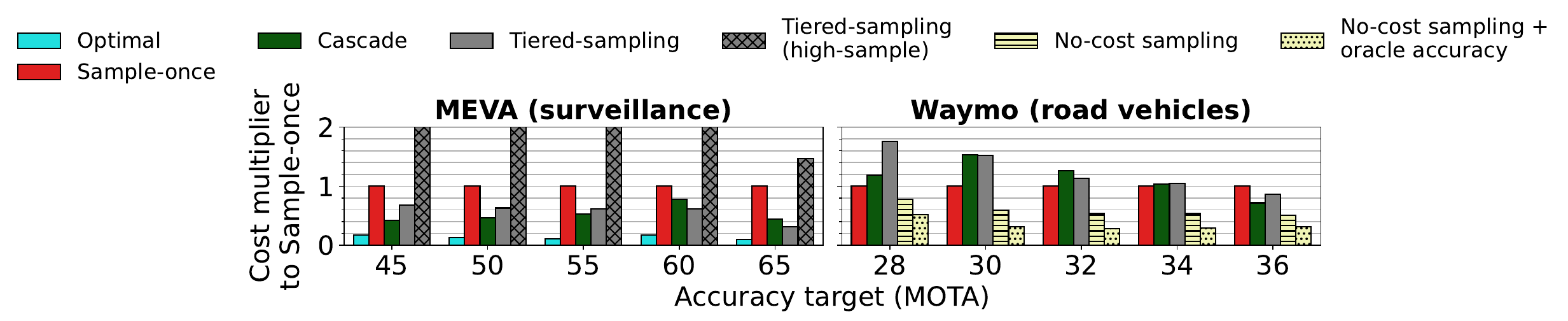}
    \caption{\small \textbf{The cost of various model size selection methods in
    different video workload types and accuracy targets.} The cost (y-axis) of
    various model size selection methods (color, bar pattern), normalized to
    the cost of the sample-once method (red) in surveillance (left panel) and
    road vehicle camera (right panel) video workloads across several accuracy
    targets (x-axis). Bars with patterns (e.g. Tiered-sampling (high-sample)),
    represent hypothetical solutions. }
    \label{f:chameleon-profile-once}
    \vspace{-1em}
\end{figure*}

\myparagraph{Determining the optimal model size} To achieve optimal \DMSS{}, it
is necessary to know both the cost and accuracy in each video segment. For
example, \Cref{f:model-accs} (top) shows the accuracies of
EfficientDet~\cite{efficientdet} models of varying sizes in a video. The
accuracies are computed over 2-second segments along the video. By knowing the
accuracy of different model sizes, the system can determine the best model
sizes in each video segment that minimize total cost while meeting the
aggregate accuracy target (\cref{f:model-accs}, bottom). Over many videos,
optimal dynamic model size selection (\Cref{f:chameleon-profile-once} left,
blue) can reduce cost by up to 10$\times$ when compared to the static selection
(``sample once''), which selects and fixes the same model size for the entire
video.

\textbf{Prediction statistics.} To overcome the absence of ground-truth labels
at runtime, prior works use \textit{prediction statistics} calculated from
model
predictions~\cite{noscope,cascade,posthoc-models,dds,EAAR,edgeduet,videostorm,chameleon,ekya,recl}.
These prediction statistics typically require only a few arithmetic operations
on model predictions (e.g., summing objectness scores~\cite{dds,edgeduet,EAAR})
and are therefore inexpensive to compute.
These works use the prediction statistics as indicators for the accuracy in
order to select the model size to run on the video segment. This is done in
many settings, including perception tasks~\cite{boggart,noscope,dds}, as well
as other
modalities~\cite{predictable_AI,karbasi2022asr,predictaboard,liang2025lumos}.


\myparagraph{Using historical workload data to set sampling/selection
parameters} The use of prediction statistics often introduces system parameters
(e.g. the confidence score threshold of model cascade~\cite{noscope,tabi,dds})
that control how models are sampled and selected~\cite{noscope,chameleon,recl}.
Determining the parameter values analytically is complex. Therefore, existing
works~\cite{dds,chameleon,recl,ekya,noscope} rely on historical workload data
that is representative of the video content encountered at runtime. This data
is used to search for the best set of parameter values that allows the system
to optimize the cost and accuracy.






\paragraph{Challenges in sampling.} Sampling too many models or none at all is
suboptimal. For example, we find (\Cref{f:chameleon-profile-once}, left) that
sampling every model in every segment (checkered gray bar, representing a
high-sampling version of one \DMSS{} method, Tiered-sampling~\cite{chameleon})
drives cost above the static selection (red ``sample-once'' bar). At the other
extreme, never sampling throughout the video results in performance equal to
that of the static selection, losing the potential savings of \DMSS{}. The
challenge is therefore to strike a balance.

In addition, we find that the optimal number of models to sample varies across
video segments and videos, as well as across workload types and accuracy
targets (\Cref{f:optimal-sampling-subset},
\cref{app:optimal-subset-sizes-other-settings}).


\myparagraph{Challenges in selection} After a subset of the models is sampled,
first, their prediction statistics are implicitly or explicitly used to
estimate the accuracy, and then the estimated accuracy is used to select the
best model size. However, as shown in \Cref{f:pred-stats-acc-corr}, the
prediction statistics often only have a weak correlation (e.g., $<0.4$) with
the accuracy. It is very uncommon that the correlation is strong (e.g.,
$>0.8$), which would indicate a direct, linear relationship between the
prediction statistics and the accuracy. These results suggest that the
relationship between the prediction statistics and the accuracy is often weak
and potentially noisy. Since the best model size to select is determined from
the accuracy, the lack of a strong linear relationship between the prediction
statistics and the accuracy makes it challenging to determine the best model
size from the prediction statistics.

\setcounter{section}{2}
\section{\system{}'s Solution approach}
\label{s:solution-approach}

We first describe the problem with existing works that use manually-crafted
rules for sampling and selection
(\cref{ss:poor-performance-mobile-low-accuracy}). Then, we present our ideas to
improve sampling (\cref{ss:improving-sampling}) and selection
(\cref{ss:improving-selection}), which guide the design of \system{}.

\subsection{Problem with existing works}
\label{ss:poor-performance-mobile-low-accuracy}
Prior \DMSS{} methods \cite{chameleon,noscope,dds,remix} decide which model to
run by using manually-crafted rules for \emph{sampling} and \emph{selection}.
These rules were developed for static-camera workloads (e.g., surveillance) and
for high accuracy targets. As expected, on surveillance videos these methods
(e.g., cascade~\cite{noscope,dds,tabi,tahoma},
Tiered-sampling~\cite{chameleon}) beat the simple baseline that selects and
fixes one model size for the rest of the video (\emph{sample-once}); see
\Cref{f:chameleon-profile-once}, left. However, when we adapt the same methods
to mobile-camera workloads (e.g., road vehicle video) and to lower accuracy
targets, the results flip: they cost \emph{more} than the sample-once baseline
in low-to-medium accuracy targets (\cref{f:chameleon-profile-once}, right).

To understand the cause, we separate the effect of sampling from the effect of
selection using two diagnostic variants:

\noindent\textbf{(a) Inefficient sampling.} We first remove the cost of
sampling from existing works (Tiered-sampling) while keeping the original
selection rules (\emph{no-cost sampling}, \cref{f:chameleon-profile-once}
right). The remaining gap to ``sample-once'' reflects the cost savings achieved
by \DMSS{} alone. On road vehicle videos, this gap is small (e.g., $\sim$20\%
cost reduction at 28 MOTA, \cref{f:chameleon-profile-once} right). This gap is
\underline{less} than the cost of sampling that was removed. Together, the
models that these methods sample often do not ``pay for themselves,'' so total
cost increases.

\noindent\textbf{(b) Suboptimal selection.} Next, we also replace the original
selection rules with oracle selection, which has access to the true accuracy
that each model would achieve on the segment (\emph{no-cost sampling + oracle
accuracy}, \cref{f:chameleon-profile-once}, right). This hypothetical variant
cuts cost by as much as 50\% relative to no-cost sampling, indicating that the
accuracy prediction used by current selection rules to select a model size has
too much error, resulting in suboptimal model size selection.


\subsection{Improving sampling}
\label{ss:improving-sampling}


\myparagraph{Analysis of sampling} Sampling a model $M$ has two opposing
effects:
\begin{tightenumerate}
    \item The \textbf{cost} of running $M$ on the exemplar frames (to get its
    predictions and calculate the prediction statistic).
    \item The potential \textbf{benefit} from changing to a better model size
    selection given the calculated value of the prediction statistic of model
    $M$ (e.g. reduction in cost of the selected model for the same accuracy).
\end{tightenumerate}

For clarity, we will refer to these as \underline{component (i)} and
\underline{component (ii)}, respectively.

\noindent \textbf{Sampling should happen when the net effect is positive}, i.e.
component (ii) > component (i). Existing
works~\cite{chameleon,noscope,remix,tabi,dds} do not explicitly estimate these
components. Instead, they use rules that incorporate fixed assumptions about
their relative effect. We propose to explicitly estimate and compare these
components at runtime.

\myparagraph{Estimating the benefit (component (ii))}
Component (i) is known upfront, before sampling, as a fixed cost of model
inference. In contrast, component (ii) is the benefit we might get
\textit{after} sampling, when model $M$'s prediction statistic causes the
system to switch to more optimal model size. Since the prediction statistic is
only revealed after sampling, we cannot know the exact benefit at decision
time. To overcome this, we \emph{estimate} the benefit in advance from
historical workload data.



\begin{figure}[t!]
    \centering
    \captionsetup[subfigure]{width=0.99\textwidth}
    \includegraphics[width=0.49\textwidth]{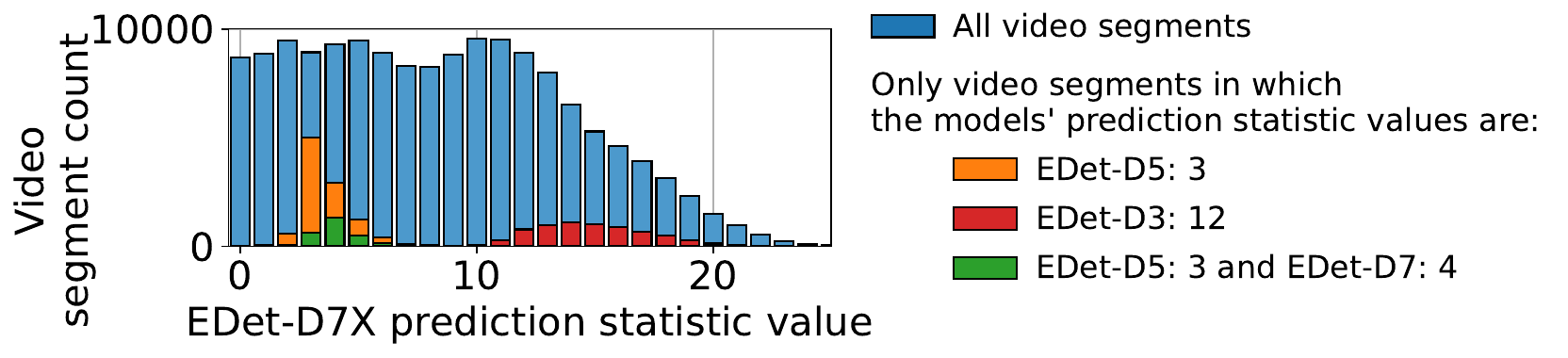}
    \caption{\small \textbf{Using the historical workload data (filtered and
    unfiltered) to get the range of likely values of prediction statistics.}
    The histogram of the prediction statistic (sum of objectness
    score~\protect\cite{dds,EAAR}) of model
    EDet-D7X~\protect\cite{efficientdet} in different video segments of the
    historical workload data (road vehicles videos~\protect\cite{waymo}). The
    histogram is shown unfiltered for all video segments (blue), and filtered:
    only for segments where EDet-D5=3 (orange), only for segments where
    EDet-D3=12 (red), and only for segments where EDet-D5=3 and also EDet-D7=4
    (green).}
    \label{f:trial-model-pred-distribution}
    \vspace{-1em}
\end{figure}

\myparagraph{Using historical workload data} Although model $M$'s prediction
statistic is unknown before sampling, its \emph{distribution} can be estimated
from historical workload data. Using empirical frequencies of past video
segments (\Cref{f:trial-model-pred-distribution}, blue), we assign
probabilities to each possible value and compute the \textit{expected value} of
the benefit (component (ii)). This expected value is compared against the known
sampling cost (component (i)) to decide whether to sample model $M$.

\begin{figure}
\includegraphics[width=\linewidth]{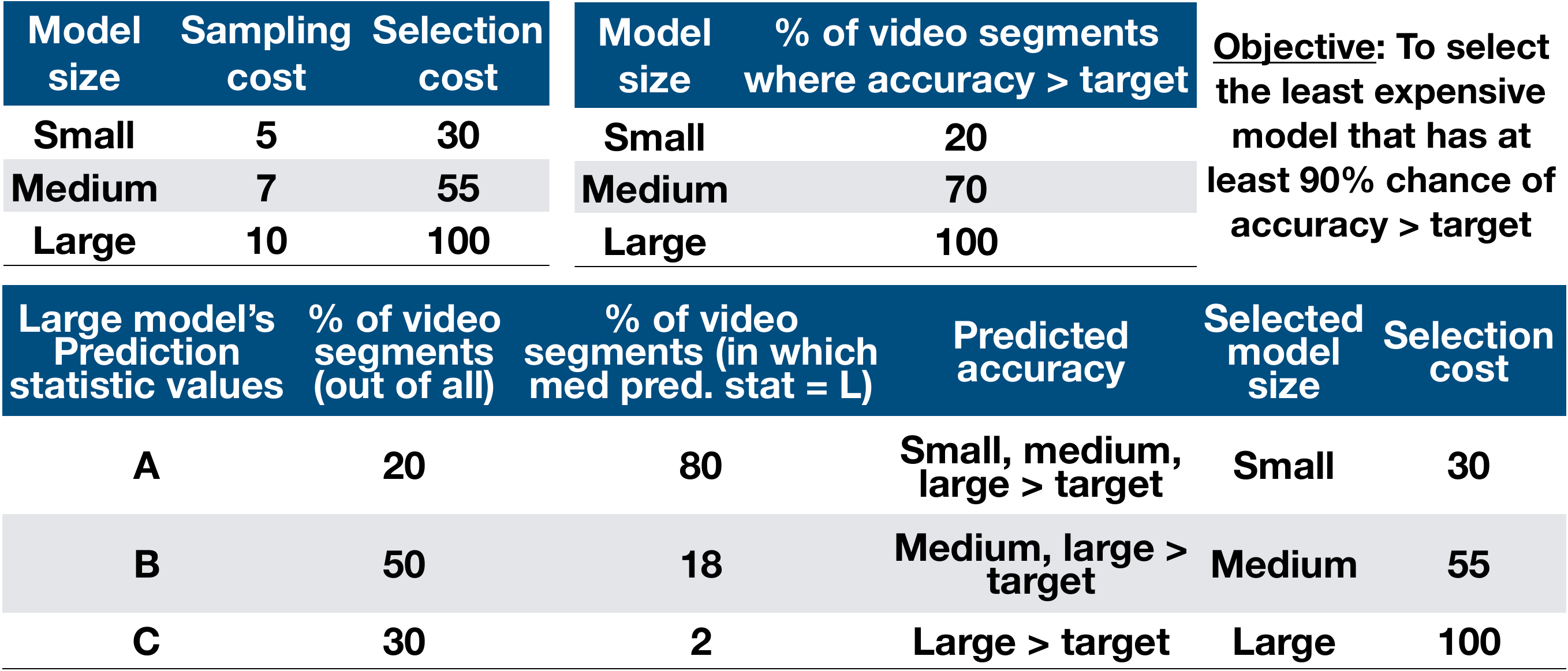}
\captionof{figure}{\small \textbf{Example for the use of historical workload
data to estimate components (i) and (ii).} Top left: The costs of sampling and
selection. Top center: Predicted accuracy when no models are sampled. Bottom:
Varying outcomes of sampling Large, based on the prediction statistic value in
the video segment. See text for details.}
\label{f:toy-example-historical-workload-data}
\vspace{-1em}
\end{figure}

\noindent \textbf{Example 1.} Consider a system with three model sizes (Small,
Medium, and Large), with selection costs of 30, 55, and 100, respectively
(\Cref{f:toy-example-historical-workload-data}). The objective (as used in
prior work~\cite{chameleon,noscope}) is to select the least expensive model
size whose accuracy has at least 90\% chance of being above the accuracy
target. We consider two possible sampling decisions and estimate component (i)
and (ii) to decide which is better:

\textbf{(a) No sampling (baseline).} Historical workload data indicates
(\cref{f:toy-example-historical-workload-data}, top center) that the Small,
Medium, and Large model's accuracies exceed the target in 20\%, 70\%, and 100\%
of the video segments. Without sampling, Large must be selected (cost=100).

\textbf{(b) Sampling the Large model.} We sample Large (cost=10), and calculate
its prediction statistic, the sum of objectness score across all objects in the
frame~\cite{dds,EAAR}. A high score indicates that the video segment's scene is
busy, and thus more challenging: More chance for small objects, occlusions, or
unusual angles that are harder to detect. If the Large model is sampled, its
prediction statistic has three possible values: A, B, and C (representing a
low, medium, and high objectness score). Historical data show
(\Cref{f:toy-example-historical-workload-data}, bottom):
\begin{tightenumerate}
    \item If the prediction statistic is A: All models meet target
    $\rightarrow$ select Small (cost=30)
    \item If B: Medium and Large meet target $\rightarrow$ select Medium
    (cost=55)
    \item If C: Only Large meets target $\rightarrow$ select Large (cost=100)
\end{tightenumerate}
The \emph{expected value} of the selection cost after sampling = $0.2 \times 30
+ 0.5 \times 55 + 0.3 \times 100 = 63.5$. The benefit from improving selection
(component (ii)) $=100-63.5=36.5$, which exceeds the sampling cost (component
(i))=10. Therefore, sampling the Large model is preferred over no sampling.

\myparagraph{Using correlations between prediction statistics} Before sampling,
we only have a broad range for a model's prediction statistic based on all past
segments (\Cref{f:trial-model-pred-distribution}, blue), making our benefit
estimate (component (ii)) imprecise. To narrow down this range of potential
values, we observe that the prediction statistics of different model sizes show
a moderate to strong correlation (0.6-0.9) across workload types, prediction
statistic types, and model families (\cref{app:pred-stats-pred-stats-corr}).
This means that prediction statistic values tend to move together, so the value
of one model's prediction statistic helps narrow down the likely values of
other prediction statistics of models that weren't sampled yet, improving
component (ii)'s estimate.

This is shown in \cref{f:trial-model-pred-distribution}: The blue histogram
shows model EDet-D7X's prediction statistic distribution across all video
segments. When the historical data is filtered only to segments where EDet-D5's
statistic=3, the likely range of EDet-D7X's prediction statistic narrows
markedly. Multiple sampled models compound this effect: Filtering by both
EDet-D5=3 and EDet-D7=4 (green) narrows the likely values further.


\textbf{Example 2: A narrower value range flips the sampling decision.}
Suppose we already sampled the Medium model and its prediction statistic is
$L$. With this observation, the tentative selection is the Medium (meets
$>$90\% objective) with cost 55. Now, consider three options:
\begin{tightenumerate}
    \item \textbf{No further sampling (baseline).} Keep Medium selection,
    cost=55
    \item \textbf{Sample Large using overall frequencies (ignoring L).} From
    historical data (\cref{f:toy-example-historical-workload-data}, bottom),
    the Large models' prediction statistic values A, B, C occur in 20\%, 50\%,
    and 30\% of the video segments. Following the same analysis in the previous
    example: The sampling cost (component (i)) = 10, the expected selection
    cost = 63.5, so the change in cost from updated selection (Component (ii))
    $= 55-63.5 = -8.5$. Component (ii) $<$ component (i) $\rightarrow$
    \underline{Don't sample.}
    \item \textbf{Sample the Large model using the filtered frequencies (taking
    L into account).} Historical data filtered by L
    (\cref{f:toy-example-historical-workload-data} (bottom)) shows that A, B, C
    occur in 80\%, 18\%, and 2\% of the video segments (the frequencies become
    concentrated on A (80\%). The sampling cost (component (i)) is 10, but now
    the expected value of the selection cost $= 0.8 \times 30+0.18 \times
    55+0.02 \times 100 = 35.9$, so component (ii) $= 55-35.9 = 19.1$. Since
    component (ii) > component (i) $\rightarrow$ \underline{Sample the Large
    model (decision flipped!)}.
\end{tightenumerate}

\textbf{Takeaway.} Using historical workload data (Ex. 1) and filtering it
using the prediction statistics of previously-sampled models (Ex. 2) yields
explicit (and more precise) estimates of component (ii), which can be used to
decide which model to sample and which to avoid.

\subsection{Improving selection}
\label{ss:improving-selection}

\begin{figure}
\includegraphics[width=\linewidth]{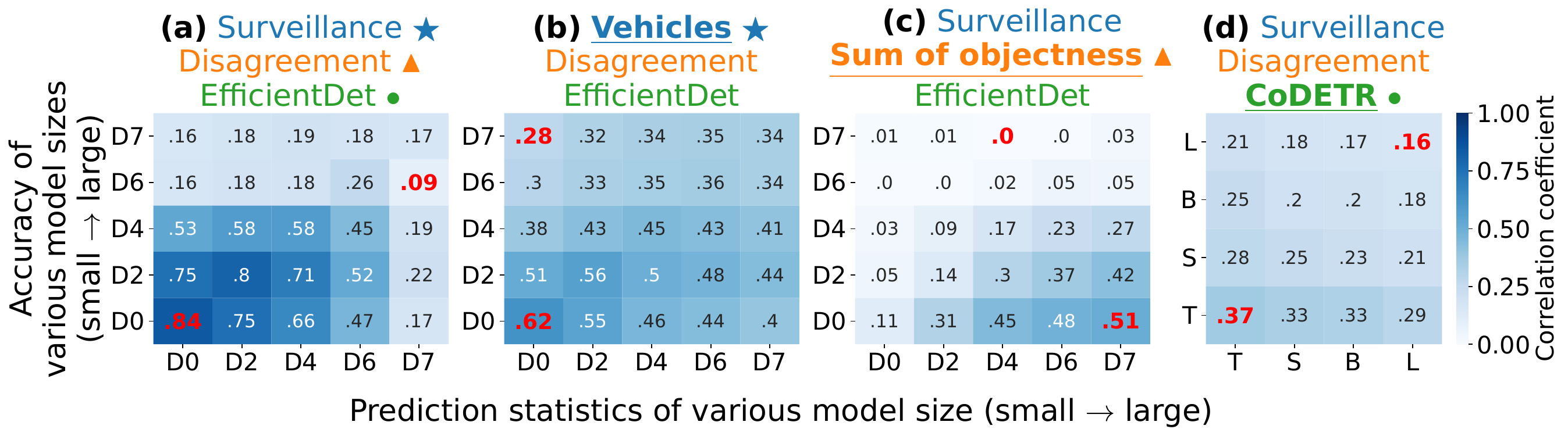}
\captionof{figure}{\small \textbf{The varied correlation between prediction
statistics and accuracy.} The correlation (Pearson) between the prediction
statistics (x-axis) and the accuracy (y-axis) across different model sizes.
Heading: \textcolor{tabblue}{Blue: Workload type}.
\textcolor{taborange}{Orange: Prediction statistic type}.
\textcolor{tabgreen}{Green: Model type}. The correlation varies depending on
the workload type (\textbf{(a)} vs \textbf{(b)}), prediction statistic
(\textbf{(a)} vs \textbf{(c)}), and model type (\textbf{(a)} vs \textbf{(d)}).
The maximum and minimum values in each heatmap are emphasized (bold/red).}
\label{f:pred-stats-acc-corr}
\vspace{-1em}
\end{figure}

\myparagraph{Varying and weak correlation between prediction statistics and accuracy}
Selection depends on predicting each model's accuracy from prediction
statistics. These statistics (e.g. prediction confidence score) are imperfect
\textit{proxy metrics} for accuracy. \Cref{f:pred-stats-acc-corr} shows that
the Pearson correlation between each model size's prediction statistic (x-axis)
and the accuracy of other models (y-axis) ranges from nearly 0 to
$\approx$0.8, and varies substantially with workload type, prediction
statistic, and model family (similar trends using Spearman correlation are
shown in \cref{app:pred-stats-acc-corr-spearman}).\footnote{More broadly, prior
work shows the relationship between common prediction statistics and accuracy
can be non-linear and dataset- and
model-dependent~\cite{ovadia2019can,platanios2016estimating,guo2017calibration}.}
This variation implies that simple hand-crafted rules (e.g., single-variable
thresholds or formulas~\cite{chameleon,tabi,dds,cascade,noscope}) will not
predict the accuracy well across all settings, motivating a more robust
approach.

\textbf{Adopting a performance modeling perspective.} We view predicting
accuracy from prediction statistics as a \textit{performance modeling} problem.
In systems scheduling, performance models are widely used when the mapping from
task configuration and proxy metrics (e.g., \% CPU utilization) to the
end-to-end metric (e.g., latency) is complex and
platform-dependent~\cite{cherrypick,earnest,paragon,quasar,sinan,jockey,cilantro,linnos}.
Our setting is analogous: the mapping from prediction statistics to accuracy
depends on complex interactions between the model and the input distribution
and can vary across models and workload types.

\textbf{A data-driven performance model for accuracy.} We therefore use a
learned performance model~\cite{sinan,paragon,quasar} that leverages historical
workload data to predict the accuracy of each model size from prediction
statistics. Compared with manually-crafted rules, a learned model can combine
multiple weak proxy metrics and remain robust in the presence of
noise~\cite{linnos,cilantro}. In \cref{ss:eval-ablation}, we show this
data-driven approach improves accuracy prediction substantially
($3.5\times$--$10\times$ lower error versus manually-crafted baselines), which
in turn improves model selection.

These key insights (explicit cost-benefit analysis for sampling and using a
data-driven accuracy performance model for selection) guide \system{}'s design.

\section{\system{} Design}
\label{s:design}

\begin{figure}[t!]
    \centering
    \captionsetup[subfigure]{width=0.99\textwidth}
    \includegraphics[width=0.48\textwidth]{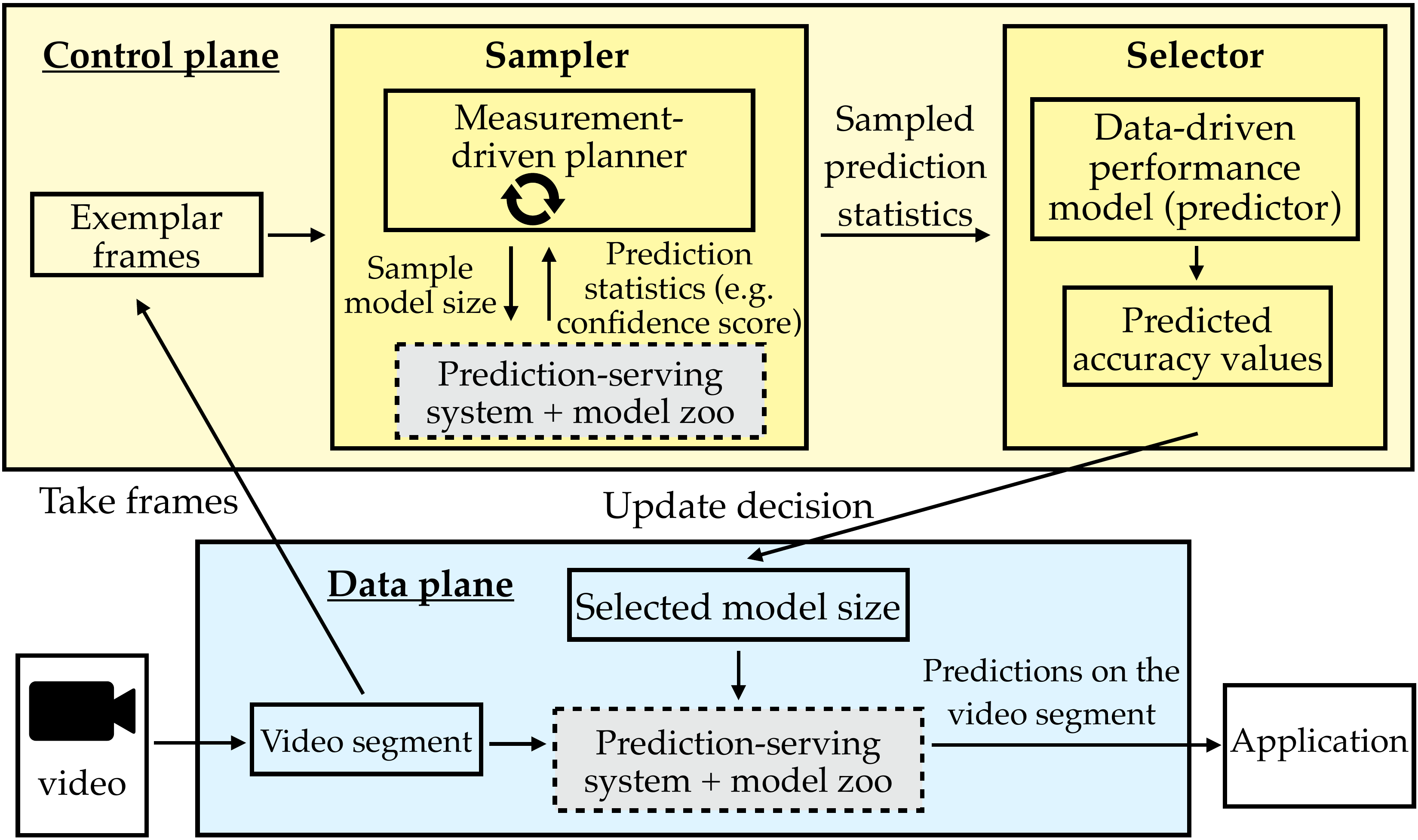}
    \caption{\small \textbf{\system{}'s architecture and workflow.} }
    \label{f:system-arch-workflow}
    \vspace{-1em}
\end{figure}

\system{} is a dynamic model size selection system for live video analytics.
Its goals are \one{} simplicity: no changes to the base-models or additional
training, and \two{} cost reduction by adapting model size to video content.
\system{} works with any prediction-serving system and model zoo
(\Cref{f:system-arch-workflow}). It consists of a control plane for sampling
and selection and a data plane that executes the selected model size. The
control and data planes are separated to keep selection off the critical path
(\cref{ss:design-control-data-separation}) and give the data plane priority
when selection and execution call the same prediction-serving system.

\myparagraph{\system{}'s workflow (\Cref{f:system-arch-workflow})} For each new
video segment, exemplar frames are sent to the control plane. Inside the
sampler (\cref{ss:design-sampler}), a measurement-driven planner uses
historical workload data to guide an iterative sampling process that
efficiently determines which model to sample or avoid. The prediction
statistics of the models sampled are passed to the selector
(\cref{ss:design-selector}), which uses a data-driven performance model to
predict each model's accuracy. \system{} then calculates the cost-accuracy
objective (\cref{ss:design-objective}) for each model, selects the best model
size, and sends the selection back to the data plane to run on the video
segment.

\subsection{Objective}
\label{ss:design-objective}
The application's cost and accuracy goals are calculated in aggregate over all
videos at the end of execution. In contrast, at runtime, LVA processing only
has access to past videos. Therefore, different works
\cite{videostorm,chameleon,noscope} use a surrogate objective to guide sampling
and selection in each video segment.

\system{} uses a scalarized objective that linearly combines normalized
accuracy and cost in a single expression, using a tunable trade-off parameter
$\alpha\in[0, 1]$:

{
\setlength{\abovedisplayskip}{2pt}
\setlength{\belowdisplayskip}{2pt}
\begin{align}
    \label{e:cost-accuracy}
    \text{cost}\_\text{acc}\_\text{obj} = \alpha \times \text{cost} + (1 - \alpha) \times (1 - \text{acc})
\end{align}
}

This objective is simple to calculate and is used extensively in utility and
objective functions in prior work~\cite{videostorm,deepdecision,vulcan}.

\subsection{Efficient Sampling}
\label{ss:design-sampler}

\cref{s:solution-approach} shows how historical workload data can be used to
estimate component (ii), and that as more prediction statistics are known, the
estimates and the sampling decisions improve.

\textbf{Sequential operation.} To maximally exploit these ideas, \system{}
avoids sampling models in parallel, which precludes using one model's
prediction statistic to improve sampling for other models. Instead, we design a
measurement-driven planner that samples sequentially, one model at a time. This
allows using the observed prediction statistics of all previously sampled
models in order to get increasingly precise estimates of component (ii) for the
next model to sample.

This sequential sampling design abides by anytime semantics: As more iterations
are run, more cost is expected to be saved since each additional iteration is
designed to improve the objective through careful estimation of component (ii).


\myparagraph{Probability estimation from historical workload data}
Component (ii) is estimated using the expected value of the prediction
statistic of the model $m$ considered for sampling. This requires estimating
$p_m(s|\mathcal{O})$, the probability that prediction statistic value $s$ will
be calculated after sampling model $m$ on the video segment, given prediction
statistics set $\mathcal{O}$ observed from previously sampled models. Since the
sampling process may repeatedly query $p_m(s|\mathcal{O})$ to calculate the
expected value, its estimation must be computationally inexpensive and fast.
Therefore, we use empirical probability estimates for $p_m(s|\mathcal{O})$
based on observed frequencies in the historical
data~\cite{wasserman2013all_stats_textbook_1,manning2008introduction_stats_textbook_2,agresti2025categorical_stats_textbook_3}.

Let $\mathcal{O}=\{(m_i,s_i)\}_{i=1}^k$, where prediction statistic $s_i$ was
observed for model $m_i$ in the current video segment. Define $N(\mathcal{O})$
to be the number of video segments in the historical data in which these
prediction statistics were also observed:

\begin{equation}\label{e:probability-estimation}
\widehat{p}_m(s \mid \mathcal{O}) \;=\; \frac{N(\mathcal{O} \cup \{(m, s)\})}{N(\mathcal{O})}
\end{equation}


Continuous prediction statistics (e.g. confidence score~\cite{tabi}, sum of
objectness score~\cite{dds}) are discretized via binning. The number of bins is
tuned for each workload type based on the historical data. We found that using
10-20 bins for each prediction statistic is a good default across workload
types.

\textbf{Sampling planner overview.} The planner operates iteratively:
\textbf{(1)} Estimate component (ii) and the net benefit (component (ii) -
component (i)) for each unsampled model, \textbf{(2)} Sample the model with the
highest net benefit, stopping when no models have positive net benefit. To
estimate component (ii), the planner calls the selector and the accuracy
prediction model (\cref{ss:design-selector}) for each potential prediction
statistic value.

\textbf{Estimating component (ii).} We define $J(\mathcal{O})$ as the objective
when selecting based on the prediction statistics in $\mathcal{O}$: {
\setlength{\abovedisplayskip}{2pt}
\setlength{\belowdisplayskip}{2pt}
\begin{equation}
\begin{split}
    J(\mathcal{O}) = Objective(&c(Select(\mathcal{O})),\ \widehat{a}(Select(\mathcal{O}), \mathcal{O}))
\end{split}
\end{equation}
} $Select(\mathcal{O})$ is the model selected by the system based on prediction
statistics $\mathcal{O}$ (\cref{ss:design-selector}), $c(\cdot)$ is the model's
selection cost, $\widehat{a}$ is the predicted accuracy of the model
(\cref{ss:design-selector}), and $Objective$ is the objective function
(\cref{e:cost-accuracy}).

Component (ii) is then estimated as:

{
\setlength{\abovedisplayskip}{2pt}
\setlength{\belowdisplayskip}{2pt}
\begin{equation}\label{e:component-ii}
\begin{split}
    &Component\ ii(m, \mathcal{O}) = \\
    &\mathbb{E} [J(\text{pred. stats before sampling})] \\&- J(\text{pred. stats after sampling}) = \\
    &J(\mathcal{O}) - \sum_{s \in \mathcal{S}_m} \widehat{p}_m(s | \mathcal{O}) \times J(\mathcal{O} \cup \{(m, s)\})
\end{split}
\end{equation}
}

Where $\mathcal{S}_m$ is the potential values that model $m$'s prediction
statistic can take. Using \cref{e:component-ii}, the full sampling routine for
a new video segment is shown in \Cref{alg:sampling-planner}.

\begin{algorithm}[t]
\caption{Measurement-driven sampling planner}
\label{alg:sampling-planner}
\DontPrintSemicolon
\KwIn{Candidate models $\mathcal{M}$}
\KwOut{Prediction statistics $\mathcal{O}$}
$\mathcal{O} \gets \varnothing$ \tcp*{Observed (model, stat) pairs}
\While{true}{
    \ForEach{unsampled $m \in \mathcal{M}$}{
    $G[m] \gets Comp.\ ii(m, \mathcal{O}) - \textsc{SampleCost}(m)$ \tcp*{Eq.~\eqref{e:component-ii}}
    }
    \If{$\max_{m} G[m] \le 0$}{\Return $\mathcal{O}$}
    $m^\star \gets \arg\max_{m} G[m]$; \quad $s \gets \textsc{Sample}(m^\star)$; \\
    $\mathcal{O} \gets \mathcal{O} \cup \{(m^\star, s)\}$
}
\end{algorithm}



\myparagraph{Data drift and outlier detection} At runtime, sampling yields
various combinations of values of different prediction statistics
($\mathcal{O}$). To ensure we only trust probability estimates
$\widehat{p}_m(s|\mathcal{O})$ that are sufficiently supported by data,
\system{} requires at least $T_{min}$ matching video segments (by default
$T_{min}=10$) in the historical data. If the matched count falls below
$T_{min}$, the video segment's prediction statistic combination $\mathcal{O}$
is treated as underrepresented in the data. In this case, \system{} halts
sampling, removes the last prediction statistic from $\mathcal{O}$ whose
inclusion caused the count drop, and passes the remaining prediction statistics
to the selector. This approach allows \system{} to avoid inflating cost by
running long sampling sequences on new video data not sufficiently represented
in the historical data. In the extreme, if an outlier is detected after
sampling the first model, \system{} immediately falls back on the default model
size for $\mathcal{O} = \varnothing$. This mechanism gracefully handles severe
data drift. We extensively evaluate robustness to drift in
\cref{ss:eval-robustness}.

\subsection{Data-driven accuracy prediction}
\label{ss:design-selector}

\textbf{Selector overview.} As shown in \cref{alg:selector}, the selector
predicts the accuracy of each candidate model size and chooses the one
that minimizes the cost-accuracy objective.

\begin{algorithm}[t]
\caption{Selector}
\label{alg:selector}
\DontPrintSemicolon
\KwIn{prediction\ statistics $\mathcal{O}$, candidate models $\mathcal{M}$}
\KwOut{Selected model $m^\star$}
\ForEach{$m \in \mathcal{M}$}{
    $\hat{a}[m] \gets \textsc{AccuracyPredict}(m, \mathcal{O})$ \tcp*{\cref{ss:design-selector}}
    $obj[m] \gets Objective(c(m), \hat{a}[m])$ \tcp*{\cref{ss:design-objective}}
}
$m^\star \gets \operatorname*{arg\,min}_{m \in \mathcal{M}} obj[m]$; \Return $m^\star$
\end{algorithm}

\textbf{Accuracy performance model.} \system{} uses a learned performance model
to capture complex interactions between prediction statistics and accuracy. The
performance model must: \one{} tolerate changing input features (varying
$\mathcal{O}$ per segment), \two{} be computationally fast for repeated
invocations.

For our design, we use a k-nearest neighbors (kNN) model for accuracy
prediction~\cite{hastie2009elements,jiawei2006data}. kNN can naturally handle
missing features~\cite{jiawei2006data}, achieves sub-ms inference on a single
CPU core, and requires no training beyond storing historical data. When using
kNN for accuracy prediction, the same neighbors (computed from the prediction
statistics $\mathcal{O}$) predict accuracy for all $\mathcal{M}$ models, via
model-specific voting, which reduces the number of operations done by the
model. When no model has been sampled ($\mathcal{O}=\varnothing$), the
predictor returns each model's average accuracy in the historical data, which
results in selecting the best model on average. We tune $k$ and the distance
metric on historical data.

\textbf{Caching.} To further reduce the accuracy prediction overhead, \system{}
caches the accuracy predictions for commonly repeated combinations of
prediction statistics (e.g. the empty set $\mathcal{O} = \varnothing$, when no
models have been sampled yet).

\subsection{Managing latency}
\label{ss:design-control-data-separation}

Many video analytics applications impose deadlines of 2-15
sec~\cite{videostorm,enterprise_AID_2022}. Sequential sampling may overrun the
deadline because of transient delays in the prediction-serving system (e.g.,
from cloud contention). To ensure that the best model size selection available
is submitted on time for execution, we decouple the control plane
(sampling/selection) from the data plane (execution), giving the data plane
higher priority.

\textbf{Workflow.} On arrival at time $t_{\mathrm{arr}}$, each video segment is
assigned a default model size (the selector's choice when
$\mathcal{O}=\varnothing$) and a control-plane commit timestamp
$t_{\mathrm{commit}}$. At each planner iteration (\cref{alg:sampling-planner}),
the control plane passes the current prediction statistics $\mathcal{O}$ to the
selector, obtains a tentative choice $m_{\mathrm{curr}}$, and asynchronously
updates $\langle m_{\mathrm{curr}},\ t_{\mathrm{commit}}\rangle$. The commit
time is:

$$
t_{\mathrm{commit}}=t_{\mathrm{arr}}+T_{\mathrm{deadline}}-\hat{q}-\hat{t}_{\mathrm{exec}}(m_{\mathrm{curr}})
$$

where $\hat{q}$ is the estimated queuing delay and
$\hat{t}_{\mathrm{exec}}(m_{\mathrm{curr}})$ is estimated execution time (from
past latency profiles). At $t_{\mathrm{commit}}$, the data plane executes the
latest $m_{\mathrm{curr}}$. $\hat{q}$ is estimated via a direct measurement. We
leave exploring richer queue wait time prediction for future work.

\section{Evaluation}
\label{s:evaluation}

We evaluate \system{} on road vehicle, drone, and surveillance camera videos in
two perception tasks and three model architecture families, making the
following findings:
\begin{tightenumerate}
    \item \system{} reduces compute cost by 14-62\% over
    Tiered-sampling~\cite{chameleon} and 11-60\% over a gating
    network~\cite{recl} while maintaining the same accuracy in road vehicle
    videos.
    \item The majority (62\% on average) of \system{}'s performance improvement
    comes from the sampler.
    \item \system{} remains robust to data drift that results in a mismatch
    between the historical data and the videos at runtime.
    \item Additional sampling iterations monotonically improve performance and
    contribute to 5\%-40\% of \system{}'s overall improvement.
\end{tightenumerate}

\subsection{Experimental setup and methodology}
\label{ss:evaluation-setup}

\myparagraph{Implementation} \system{} is implemented using Ray~\cite{ray}, and
uses Ray Serve~\cite{ray_serve} as the prediction serving system. We use
separate actor pools for the control and data planes, which communicate
asynchronously (\cref{ss:design-control-data-separation}). During runtime, we
use nodes with 2 Titan RTX GPUs.

\myparagraph{Datasets}
We evaluate \system{} on road vehicle, drone, and surveillance camera datasets,
covering a range of camera movement speeds, weather conditions, and times of
day.
\begin{tightitemize}
    \item \textbf{Road Vehicles}: We use the front-facing camera footage in the
    Waymo~\cite{waymo} dataset. This dataset is composed of 798 training and
    202 validation videos, all of length 20 seconds.
    \item \textbf{Drone}: We use the VisDrone~\cite{visdrone} dataset, in which
    videos shorter than 20 seconds are discarded for uniformity (leaving an
    average video length of 37 seconds). The dataset is composed of 35 training
    videos and 14 test videos.
    \item \textbf{Stationary camera surveillance}: We use the MEVA~\cite{meva}
    dataset of indoor and outdoor wall-mounted security camera footage. The
    dataset has 28 training and 8 test videos, with video lengths range from 30
    seconds to 5 minutes.
\end{tightitemize}

\myparagraph{Models} To evaluate and compare \DMSS{} methods, we use a range of
multi-object tracking and instance segmentation model sizes. The trackers are
built using EfficientDet~\cite{efficientdet}, YOLOX~\cite{ge2021yolox}, or
DETR~\cite{zong2023detrs} models for detection followed by SORT~\cite{sort} for
tracking. These models span over two orders of magnitude of cost (1.08-410
GFLOP), and range in accuracy from 25.8 to 55.1 mAP. To stay consistent with
the methodology of prior works~\cite{videostorm,chameleon,noscope}, we use
model sizes whose cost and accuracy are Pareto optimal. The models are run
using TensorFlow~\cite{tensorflow}, and are pretrained on the COCO
dataset~\cite{coco,he2017mask}.

\myparagraph{Metrics}
We evaluate model size selection in multi-object tracking and instance
segmentation, both key tasks in video analytics~\cite{videostorm,dds}.
\textbf{Accuracy} is calculated for each task using standard metrics: the MOTA
tracking score~\cite{milan2016mot16} and the segmentation Average
Precision~\cite{coco}, respectively.
\textbf{Compute cost} is measured using floating-point operations (FLOPs) based
on reported values~\cite{efficientdet,ge2021yolox}. The \underline{final
accuracy} of a model size selection method is determined as follows: \one{}
model sizes are selected in the videos \two{} the selected models make
predictions on the videos \three{} the method's accuracy is the aggregate
accuracy of these predictions on the videos. The \underline{total cost} is the
average FLOPs/frame performed by the selected models.

\myparagraph{Trace-driven simulator} We also built a simulator to test
\system{} under a wide range of configurations, resource constraints, and
longer execution times. The simulator takes as input the cached model
predictions on the video frames, as well as the inference cost logs. This
exhaustive approach allows us to mimic our implementation's execution with
high-fidelity under different constraints.

\myparagraph{Comparing methods} Model size selection methods are compared by
measuring the minimal compute cost that each method requires in order to meet
the same accuracy target.

\myparagraph{Setting control parameters} As discussed
in~\cref{s:background-motivation}, several
methods~\cite{chameleon,cascade,tabi,recl} introduce multiple parameters that
control sampling and selection. We run standard random search with the
historical workload data to find the optimal parameter value combinations for
each method at each accuracy target.


\myparagraph{Setup details}
Following the prevailing evaluation methodology used in existing
works~\cite{reducto,recl,chameleon,noscope}, we use generated labels for the
MEVA (surveillance) dataset (which lacks human-annotated labels) via large
model prediction. To get high-fidelity predictions, we combine the
state-of-the-art detection model DETA~\cite{deta} and tracking module
Unicorn~\cite{unicorn}.


%


\subsection{End-to-end results}
\label{ss:eval-main-results}

First, we evaluate and compare \system{} against the following dynamic model
size selection methods:
\begin{tightenumerate}
    \item \textbf{Tiered-sampling}: This is a sampling method used by
    Chameleon~\cite{chameleon} in which expensive sampling is performed
    infrequently while inexpensive sampling is done more frequently
    (details on control parameters in \cref{app:hp-search-details}).
    %
    %
    To keep evaluation fair, the following complementary methods are not
    included in the solution: \one{} The hill-climbing method is not used
    because optimization takes place over a single dimension (model size).
    \two{} Further, our evaluation uses camera feeds that do not share the same
    space and time dimensions, so the spatial locality approach does not apply.

    \item \textbf{Gating network}: We adapt the ``expert'' model selection
    mechanism in RECL~\cite{recl} to model size selection. In this approach, a
    lightweight gating neural network is trained to select the top-k model
    sizes on a video frame. Afterwards, runtime profiling is used to filter
    down to the top-1 model size. We adapt RECL's labeling schema by setting
    the labels to be the model size choice that best optimizes the
    cost-accuracy objective. The gating network is then trained in separate
    trials using labels produced with different threshold values to modulate
    between cost and accuracy. We use the same routing network architecture as
    RECL (a ResNet18~\cite{he2016deep}). This model size has negligible compute
    cost.
    \item \textbf{Cascade}:
    The model cascade samples models from small to
    large~\cite{tabi,cascade,noscope}. The model cascade is assembled using the
    smallest and the largest models used in each dataset
    (\cref{ss:evaluation-setup}), following existing works.
    As far as we know, there are no adaptations of a model cascade to
    multi-object tracking. Therefore, we develop a straightforward
    confidence-score based early-exit rule described in
    \cref{app:cascade-design}.
    \item \textbf{Sample-once}~\cite{remix,videostorm}: Static selection in
    which the optimal model size is determined once at the beginning of the
    video and fixed for the rest of it. To keep evaluation consistent, the
    model is determined using the same selection method used by
    Tiered-sampling~\cite{chameleon}.
\end{tightenumerate}

\begin{figure*}[t!]
    \centering
    \begin{minipage}[c]{0.75\textwidth} 
        \centering
        \includegraphics[width=\linewidth]{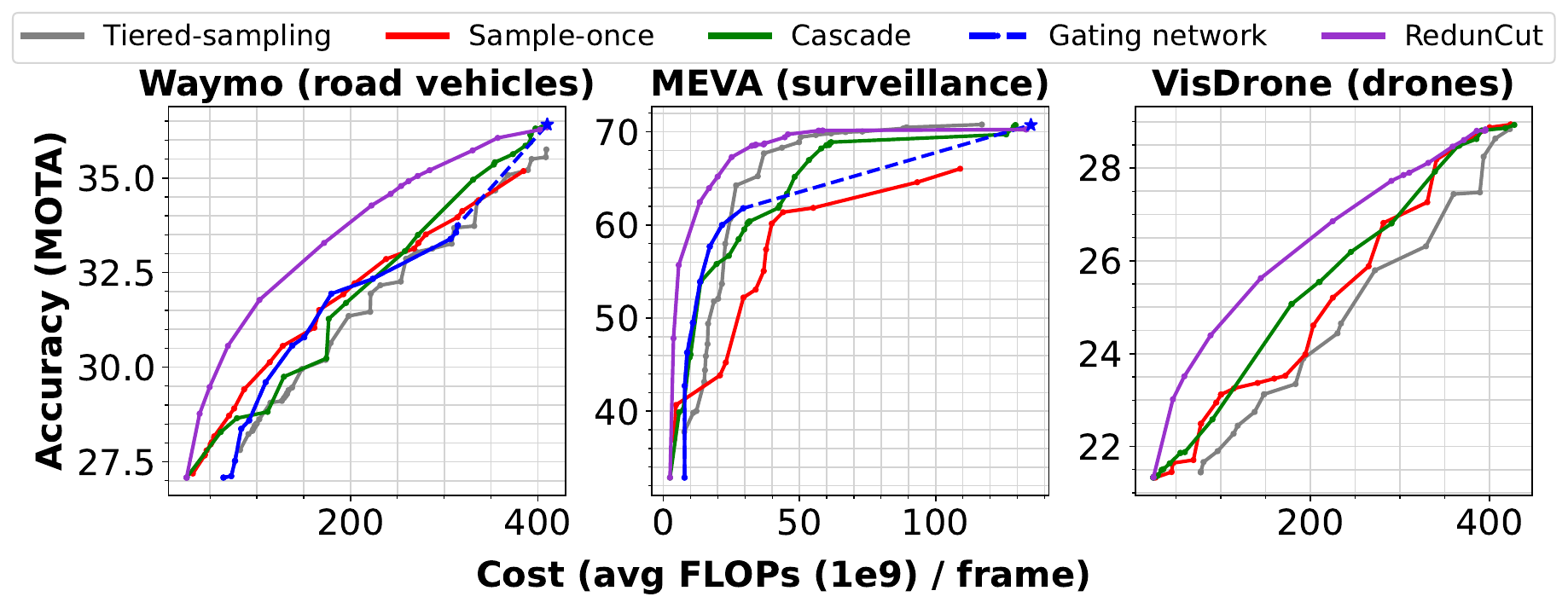}
    \end{minipage}%
    \begin{minipage}[c]{0.25\textwidth} 
        \centering
        \caption{\small \textbf{End-to-end cost-accuracy Pareto frontier of
        \system{} and other methods evaluated in road vehicle (a), Surveillance
        (b), and drone (c) videos.} The Pareto frontier of each method
        represents the lowest compute cost (x-axis) that each method requires
        to reach a certain accuracy (y-axis).
        \protect\cref{ss:evaluation-setup} describes the generation of the
        Pareto frontier curve. The dashed, blue line is described
        in~\protect\cref{ss:eval-main-results}.}
        \label{fig:main-results}
    \end{minipage}
    \vspace{-1em}
\end{figure*}


The end-to-end evaluation on surveillance, road vehicle, and drone videos is
shown in~\Cref{fig:main-results}. The figure illustrates the cost of various
methods across different accuracy targets. A line is plotted across the
cost-accuracy tradeoff points of each method to distinctly visualize them. The
results show that \system{} achieves higher accuracy at a lower cost than the
rest of the methods, across a wide accuracy spectrum and across the video
workload types. In contrast, the other methods tend to vary in their
cost-efficiency compared to one another. Existing works sample and select
models using rules that result in inefficient sampling and suboptimal
selection. Instead, \system{} achieves significant cost savings by explicitly
calculating the net effect of sampling and using data-driven accuracy
prediction.
%
%

\begin{figure}[t!]
    \centering
    \includegraphics[width=0.45\textwidth,clip]{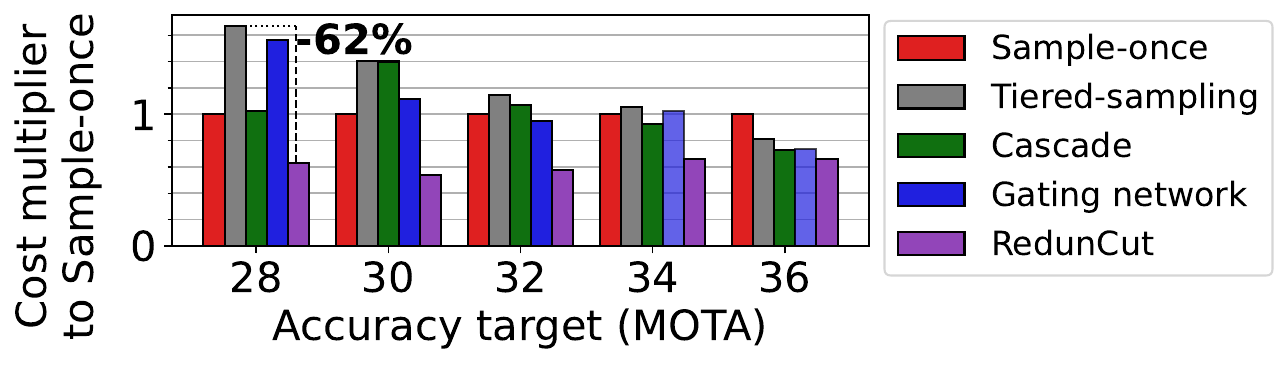}
    \caption{\small \textbf{Cost comparison between \system{} and other methods
    in road vehicle videos.} Each method (color) is evaluated on a range of
    accuracy targets (x-axis). Each method's compute cost is normalized by the
    compute cost of the sample-once method (y-axis). The translucent blue bars
    are described in~\protect\cref{ss:eval-main-results}.}
    \label{fig:barplot-main-results}
    \vspace{-1em}
\end{figure}

To more clearly compare \system{} with existing methods, we directly measure
the compute cost that each method requires in order to meet a range of accuracy
targets. To simplify comparison, the compute cost of every method is normalized
by dividing it by the compute cost of the sample-once baseline. Results are
shown on the Waymo (road vehicle) dataset in~\Cref{fig:barplot-main-results}.
We make the following observations:

\myparagraph{Comparison to \system{}} The performance difference between
\system{} and other methods varies depending on the accuracy target. In road
vehicle videos, \system{} saves 19-62\% over the Tiered-sampling
method~\cite{chameleon} and 11-60\% in compute cost over the gating
network~\cite{recl}. Likewise, in stationary camera surveillance videos,
\system{} saves 42-76\% over Tiered-sampling and 38-61\% over the gating
network (\cref{app:other-main-results}). Similar improvements are achieved in
drone videos (18-67\% over Tiered-sampling, see \cref{app:other-main-results}).

\myparagraph{Gating network limitation} Due to a limitation in the adaptation
to model size selection, the gating network's cost and accuracy do not extend
to the top-right of the tradeoff space. This is related to the large fraction
of scenarios in which smaller models have equal to or better accuracy than the
large model~\cite{kaplun2022deconstructing} (see
\cref{app:nn-prof-adaptation-limit}). To overcome this, we simulate an ideal
adaptation of the gating network in which the largest model size is always
selected. This gives the highest possible point (dashed blue segment and
asterisk in \cref{fig:main-results} and translucent blue bars
in~\cref{fig:barplot-main-results}).

 \subsection{Generalization}
\label{ss:eval-generalization}

\begin{figure}[t!]
    \centering
    \captionsetup[subfigure]{width=0.99\textwidth}
    \includegraphics[width=0.48\textwidth]{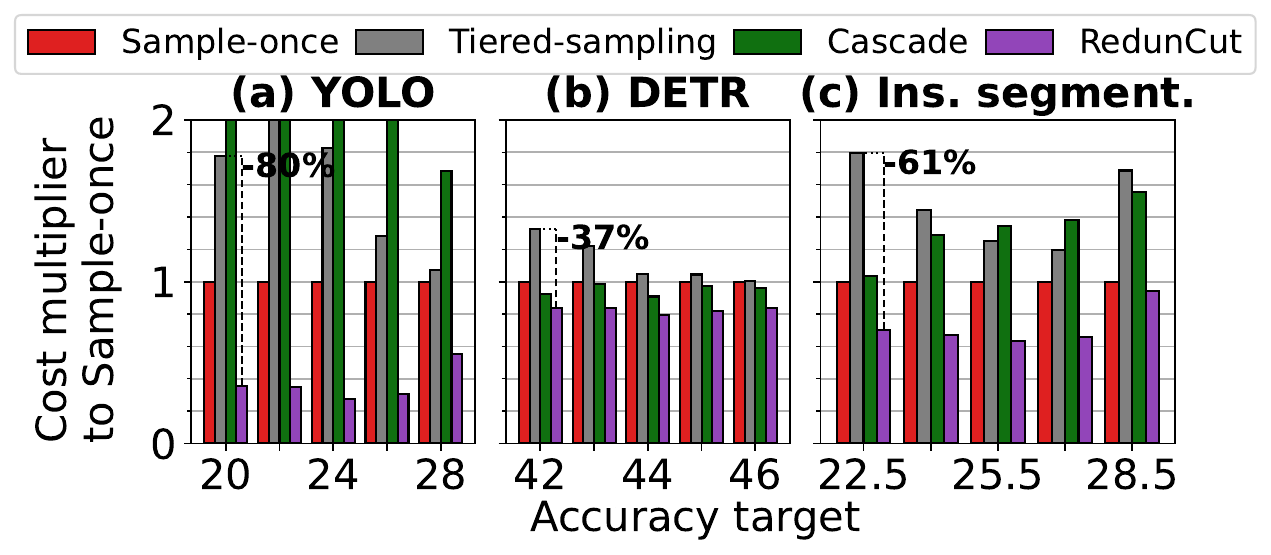}
    \caption{\small \textbf{\system{}'s performance across model families
    (a~\protect\cite{ge2021yolox}, b~\protect\cite{zong2023detrs}) and tasks
    (c, instance segmentation).} The normalized compute cost (y-axis) of
    several model size selection methods (bar color) is shown across different
    accuracy targets (x-axis).}
    \label{f:generalization}
    \vspace{-2em}
\end{figure}

\textbf{Across model families.} Next, we evaluate \system{} using different
model architecture families: The YOLOX~\cite{ge2021yolox}
(\Cref{f:generalization} (a)) and the CoDETR model
families~\cite{zong2023detrs} (\Cref{f:generalization} (b)). The cost reduction
range over Tiered-sampling is 16\%-84\% for the same accuracy target. The high
cost of the cascade baseline in the YOLOX family may be attributed to the wider
accuracy gap in the YOLOX models used in the cascade (25.8 and 49.7 mAP) than
in the EfficientDet models used in the cascade (47.2 vs 55.1 mAP). Thus, when
the accuracy gap in the cascade is higher, the larger model selection based on
the small model's confidence score is less optimal, resulting in higher cost.


\myparagraph{Across tasks} We evaluate \system{} in a different task, instance
segmentation, using the YOLOX-seg model family~\cite{ge2021yolox} and the
KITTI-STEP~\cite{weber2021step} road vehicle dataset. The results
(\Cref{f:generalization} (c)) show that \system{} improves across all accuracy
targets, with a cost reduction range of 44-61\%.


\subsection{Robustness benchmarks}
\label{ss:eval-robustness}
To further understand \system{}'s advantages and limits, we conduct evaluation
in resource-constrained settings.

\eyal{Notice that this comparison is done against baselines which have access
to the full dataset. Showing their performance in the harder data regime would
be more fair and appropriate.}

\begin{figure}[t!]
    \centering
    \captionsetup[subfigure]{width=0.99\textwidth}
    \includegraphics[width=0.49\textwidth]{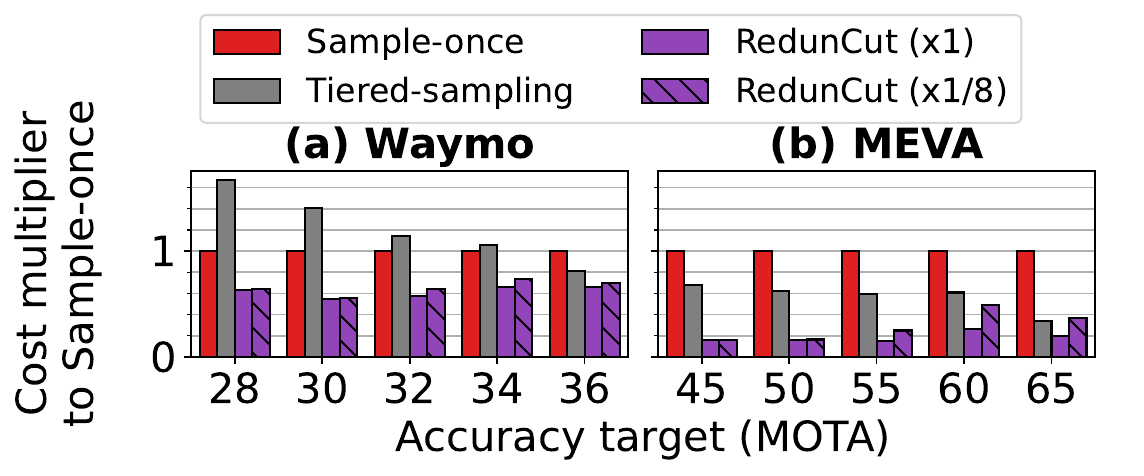}
    \caption{\small  \textbf{\system{}'s robustness in settings with less
    historical workload data.} The normalized compute cost (y-axis) of several
    model size selection policies (bar color) across different accuracy targets
    (x-axis) in road vehicles (a) and surveillance (b) videos. Historical video
    data length is reduced from 4.4 hours ($\times1$) to 0.5 hours ($\times
    1/8$) in Waymo and 2.3 hours ($\times1$) to 0.3 hours ($\times 1/8$) in
    MEVA.}
    \label{f:prep-train-fraction}
    \vspace{-1em}
\end{figure}

\myparagraph{Less historical workload data} \system {} uses the historical
workload data for both sampling and selection (\cref{s:design}). \system{} is
designed for video analytics at scale, in which large amounts of historical
data are expected to be available. However, to understand \system{}'s limits,
we investigate its performance when historical workload data is scarce. To this
end, we reduce the amount of data available to \system{} by $\times 8$
(\Cref{f:prep-train-fraction}). The results show that \system{}'s performance
is largely preserved in both Waymo and MEVA datasets (30 minutes and 18
minutes, respectively).


\begin{figure}[t!]
    \centering
    \captionsetup[subfigure]{width=0.99\textwidth}
    \includegraphics[width=0.48\textwidth]{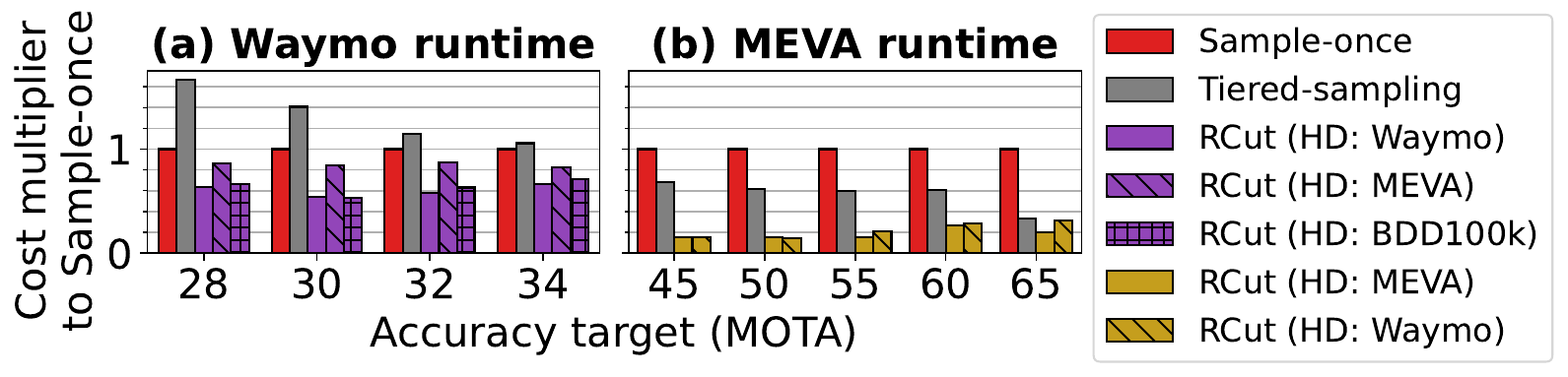}
    \caption{\textbf{\system{}'s (RCut) robustness to data drift between
    historical data (HD) and runtime videos (location, camera setup, workload
    type).} Various combinations of \system{}'s HD source were mixed and
    matched with various runtime video workloads: (a) road vehicle and (b)
    surveillance. The normalized compute cost (y-axis) is shown across
    different accuracy targets (x-axis). See
    \protect\cref{ss:eval-robustness}.}
    \label{f:prep-criss-cross}
    \vspace{-1em}
\end{figure}

\myparagraph{Robustness to drift between historical workload data and runtime
videos (location, camera setup, workload type)} Data
drift~\cite{ekya,recl,dong2024efficiently} or operational mistakes can make
runtime videos differ from \system{}'s historical data. We evaluate \system{}
by deliberately mismatching historical data and runtime videos along three
axes: geography (e.g., San Francisco/Phoenix vs. New York), camera
configuration (sensor, resolution, frame rate, mount angle), and workload type
(road vehicles vs. static surveillance). In Waymo runtime videos
(\Cref{f:prep-criss-cross}a), when we evaluate location and camera setup
mismatches (BDD100k HD, a different road-vehicle workload), cost remains within
5--10\% of using matched Waymo historical data. Mismatch in the workload type
is tougher: Using MEVA (surveillance) historical data for Waymo runtime further
increases cost, while the reverse (Waymo historical data and MEVA runtime) has
a smaller effect (\cref{f:prep-criss-cross}a--b). Together, \system{} reduces
cost over existing methods under mixed conditions, and can leverage other video
sources when matching videos are unavailable.
%
%


\begin{figure}[t!]
    \centering
    \captionsetup[subfigure]{width=0.99\textwidth}
    \includegraphics[width=0.5\textwidth]{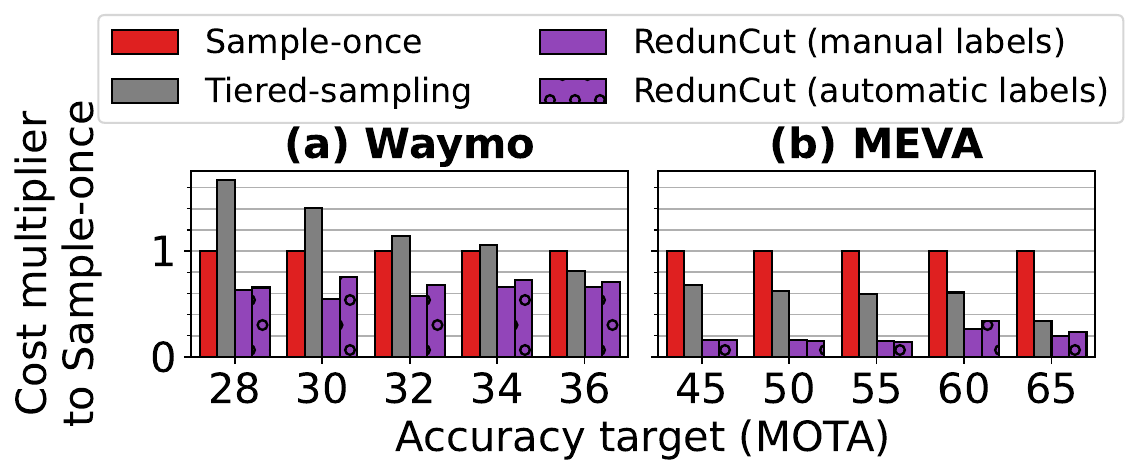}
    \caption{\small \textbf{\system{}'s robustness when the historical workload
    data is automatically labeled.} The normalized compute cost (y-axis) of
    several model size selection methods (bar color) across different accuracy
    targets (x-axis) in road vehicle (a) and surveillance (b) videos. Label
    schema: \one{} Manual human labeling (clear, purple) and \two{} Automatic
    labeling by model (purple with circles).}
    \label{f:automatic-labeling}
    \vspace{-1em}
\end{figure}

\myparagraph{Historical workload data without human labeling}
Like other methods (\cref{s:background-motivation}), \system{} requires labels
in the historical workload data. Obtaining high-quality human labels is costly
and time-consuming. A common alternative is to automatically generate labels
using a high-accuracy model, an approach adopted by several recent
works~\cite{ekya,videostorm,chameleon,recl}. Here, in
\Cref{f:automatic-labeling}, we evaluate \system{} with historical workload
data whose labels were automatically generated. The data's labels are generated
using the largest model available for selection at runtime. Results show that
across Waymo (\cref{f:automatic-labeling} (a)) and MEVA
(\cref{f:automatic-labeling} (b)), performance remains largely unchanged when
automated labels are used. This result broadens the potential use of \system{}
to settings in which human annotation is absent or expensive.


\subsection{Analysis of \system{}'s performance}
\label{ss:eval-ablation}
%

\myparagraph{The contribution of additional iterations in the planner}
In each iteration, \system{}'s planner only samples model sizes that it
estimated to have a positive net effect on cost and accuracy. To quantify the
value of additional iterations, we ablate the maximum number of iterations that
the planner can do $\{0, 1, 2, 5\}$. The 5-iteration configuration is the full
\system{} solution. With 0 iterations, no sampling is performed: \system{}
simply selects the \textit{best-on-average} model by computing each model's
average accuracy on the historical workload data and using that value as the
predicted accuracy when selecting a model size (\cref{ss:design-selector}).

\begin{figure}[t!]
    \centering
    \captionsetup[subfigure]{width=0.99\textwidth}
    \includegraphics[width=0.48\textwidth]{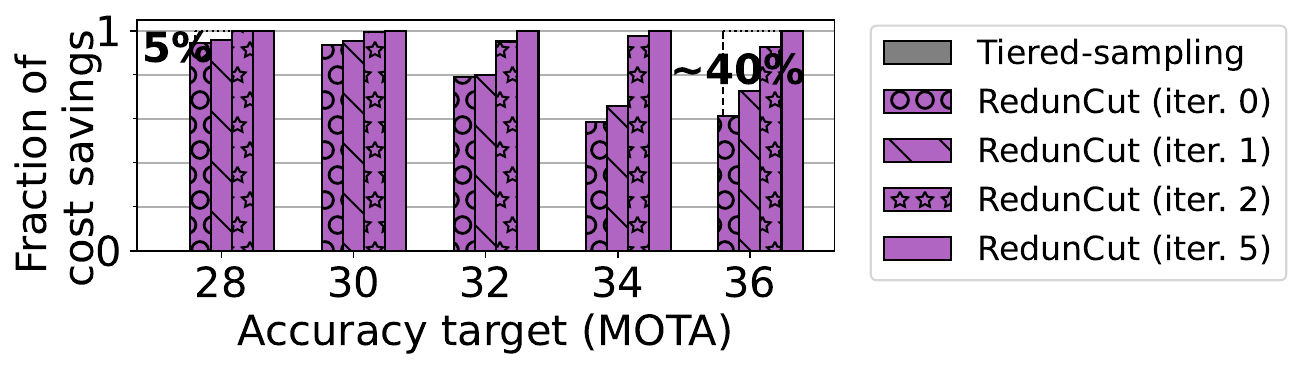}
    \caption{\small \textbf{The effect of sampling iterations on \system{}'s
    performance.} The fraction of \system's{} cost savings over
    Tiered-sampling~\protect\cite{chameleon} (y-axis) after a different number
    of sampling iterations (bar pattern) across accuracy targets (x-axis) in
    road vehicle videos. Higher y-axis values indicate better performance.}
    \label{f:ablation}
    \vspace{-1em}
\end{figure}

\Cref{f:ablation} shows that the benefit of additional iterations increases in
higher accuracy targets (x-axis). At low targets (28-30 MOTA) sampling only
adds $\sim 5\%$ of \system{}'s total cost savings, while $\ge 90\%$ of the
savings come from the best-on-average model alone. At the highest target (36)
sampling accounts for $\sim$40\% of the total savings, with the remaining
$\sim$60\% still coming from selecting the best-on-average model.

\textbf{Takeaway.} In low accuracy targets, a single model size selection using
the average accuracy suffices and additional sampling is largely redundant. As
the accuracy target increases, additional sampling iterations can help in
improving selection between small and large models.

\begin{figure}[t!]
    \centering
    \captionsetup[subfigure]{width=0.99\textwidth}
    \includegraphics[width=0.48\textwidth]{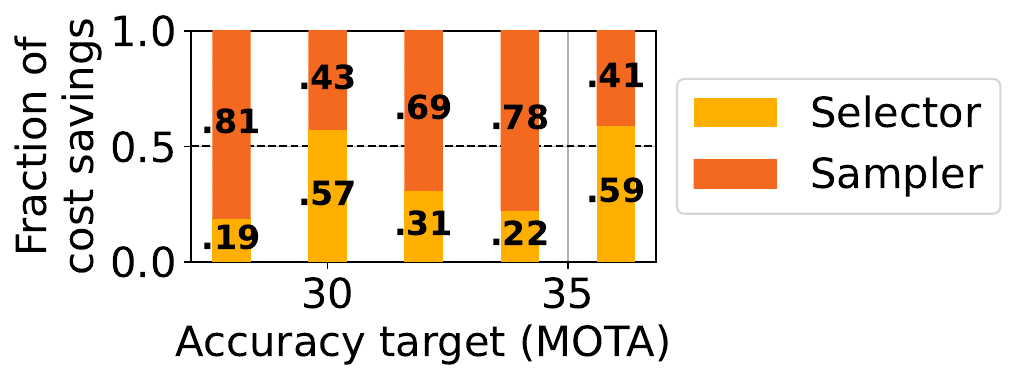}
    \caption{\small \textbf{The contribution of the sampler and selector to
    \system{}'s performance.} \system's{} cost improvement over
    Tiered-sampling~\protect\cite{chameleon} (y-axis) contributed by each
    component (color) across different accuracy targets (x-axis) in Waymo (road
    vehicle) videos.}
    \label{f:ablation-sampler-selector}
    \vspace{-1em}
\end{figure}

\myparagraph{Ablation of sampler vs selector} \system{} replaces the manually
crafted rules in prior work by using two new mechanisms: \one{} efficient,
measurement-driven sampling by the planner
(\cref{ss:improving-sampling,ss:design-sampler}) and \two{} selection via
data-driven accuracy prediction by \system{}'s selector
(\cref{ss:improving-selection,ss:design-selector}). We quantify each
mechanism's contribution by swapping our sampler with
Tiered-sampling~\cite{chameleon} while retaining \system{}'s selector. In
\Cref{f:ablation-sampler-selector}, both sampling and selection are crucial to
\system{}'s improvement over prior methods; the sampler accounts for 41--81\%
of \system{}'s total cost savings. The opposite ablation (\system{}'s sampler
and Tiered-sampling's selection), shows a similar attribution to performance by
the selector (\cref{app:selector-ablation}), suggesting a degree of overlap in
effect.

\myparagraph{Error in accuracy prediction} We evaluate \system{}'s learned
performance model (kNN) for predicting accuracy from prediction statistics, and
compare it to rules used in prior works. Across workload types (Road Vehicles,
Drones), model families (EfficientDet, YOLOX), and model sizes, \system{}
reduces prediction error by 3.5$\times$--10$\times$ vs.
Tiered-sampling~\cite{chameleon} (\Cref{f:accuracy-prediction-error};
\cref{app:accuracy-prediction-error}).

\section{Related works}


\myparagraph{Dynamic model size selection} A number of works have taken
advantage of the tradeoff between cost and accuracy across model
sizes~\cite{rafiki,figo,guo2022sommelier,cocktail,proteus}. \DMSS{} is a
narrower setup in which the system navigates this tradeoff per input on live
data. This setting is challenging since decision making is more local (lacking
access to all video data at a given point in time) and requires finer-grained
decisions using rapidly changing information. As discussed in
\cref{s:introduction} and \cref{s:background-motivation}, \DMSS{} has been
explored in both video analytics and other
applications~\cite{noscope,focus,chameleon,tahoma,tabi,remix}. We show that
these approaches fail to generalize to important workload types and accuracy
targets, and we propose \system{} to fill this gap.

\myparagraph{Complementary first-stage data similarity and deduplication
methods} Chameleon~\cite{chameleon} uses cross-camera similarity to reduce the
number of configurations that they sample. Potluck~\cite{guo2018potluck} uses a
cache-based system to detect duplicated inputs across video sources, and other
works~\cite{noscope,filterforward,reducto} explore frame skipping on the same
video when they are expected to remain same. Frame dropping is complementary to
our method, because (like recent works~\cite{ekya,recl}) we focus on resource
allocation to model inference once it is determined that the input cannot be
deduplicated or dropped.



\myparagraph{``Expert'' model training and selection}
Recent works \cite{ekya,recl,AMS}, present cost-effective methods to boost the
accuracy of a small model by fine-tuning it on specific video content at
runtime (creating ``experts'') using student-teacher
learning~\cite{hinton2015distilling}. To further save cost, RECL~\cite{recl}
proposes a ``gating network'' to select between previously trained experts.
These methods show promise in environments with limited memory and compute.
However, when compute and memory are sufficiently available, system designers
may prefer to avoid the complexity of training expert models on the fly: Model
training requires expertise, and maintaining a large, dynamic pool of models
($>150$ experts~\cite{recl}) is expensive. Further, even after training, expert
models may still substantially deviate from the predictions of the teacher
model~\cite{recl,AMS}. Instead, in environments with sufficient memory and
compute, the teacher model may be run directly.

\myparagraph{General model similarity} The use of prediction statistics is part
of a line of works that aim to quantify and characterize model
differences~\cite{jia2021zest,shah2023modeldiff,klabunde2025similarity}. The
methods in these works could be used to improve accuracy prediction at runtime.

\section{Conclusion}

Dynamic model size selection can significantly reduce LVA cost, but prior
methods use manually-crafted rules that oversample and misselect outside their
intended workload types and accuracy targets. \system{} reframes sampling as an
explicit cost-benefit decision from historical data, and predicts accuracy
using a lightweight performance model. As a model-agnostic shim layer with
sequential sampling, \system{} lowers compute cost by 14-62\% for a given
accuracy target across diverse workload types, model families, and tasks.

\newpage

\bibliographystyle{unsrt}  
\bibliography{paper}

\clearpage

\begin{appendix}
\section*{Supplementary Material: Table of contents}
\begin{itemize}
    \item Control parameter search details (\cref{app:hp-search-details})
    \item Gating network adaptation limitation (\cref{app:nn-prof-adaptation-limit})
    \item Multi-object tracker cascade design (\cref{app:cascade-design})
    \item MEVA and VisDrone main results (\cref{app:other-main-results})
    \item Spearman correlation between prediction statistics and accuracy
          (\cref{app:pred-stats-acc-corr-spearman})
    \item Optimal number of models to sample in different settings
    (\cref{app:optimal-subset-sizes-other-settings})
    \item Accuracy prediction error (\cref{app:accuracy-prediction-error})
    \item Correlations between prediction statistics (\cref{app:pred-stats-pred-stats-corr})
    \item Selector ablation (\cref{app:selector-ablation})
\end{itemize}


\section{Control parameter search details}
\label{app:hp-search-details}
\myparagraph{Tiered-sampling}
The following knobs are left as control parameters which we search and find
separately as described in \cref{ss:evaluation-setup}: profiling window length,
segment length, subsegment profiling duration, the configuration subset size
$k$, and the size of the reference model used as ground truth when sampling the
top-$k$.

\section{Gating network adaptation limitation}
\label{app:nn-prof-adaptation-limit}
\textbf{Why doesn't the gating network reach the top-right of the tradeoff space?}
This happens because of the way that the gating neural network in
RECL~\cite{recl} is adapted to model size selection in our setting. The
adaptation trains the gating network to select the model size that optimizes
the cost-accuracy objective used in \system{}. The downside of this approach is
that when the objective is set to prioritize maximizing accuracy (with no limit
on cost), 49\% of the model size labels (in the Waymo dataset) are not the
largest model size. This takes place in scenarios in which a small model has an
accuracy equal to or (occasionally) greater than the accuracy of the large
model~\cite{kaplun2022deconstructing}. As a result, the gating network that is
adapted for model size selection is never trained to select the largest model
for all video segments.

\section{Multi-object tracker cascade design}
\label{app:cascade-design}
To the best of our knowledge, there are no known works that have adapted a
model cascade for multi-object tracking. Therefore, we develop the following
confidence-score based early-exit rule: the smaller model is run on the frame,
and the resulting object predictions with a confidence score in the range [$l$,
$h$) are recorded by the cascade. Then, their average confidence score $m$ is
computed and compared against a confidence threshold $c$. When $m < c$, the
larger model is run next and its predictions are chosen for the frame.
Otherwise, the larger model is not run and the predictions of the smaller model
are chosen for the frame. $l$, $h$, and $c$ are left as control parameters that
are searched as described in \cref{ss:evaluation-setup}.

\section{MEVA and VisDrone main results}
\begin{figure}[t!]
    \centering
    \includegraphics[width=0.45\textwidth,clip]{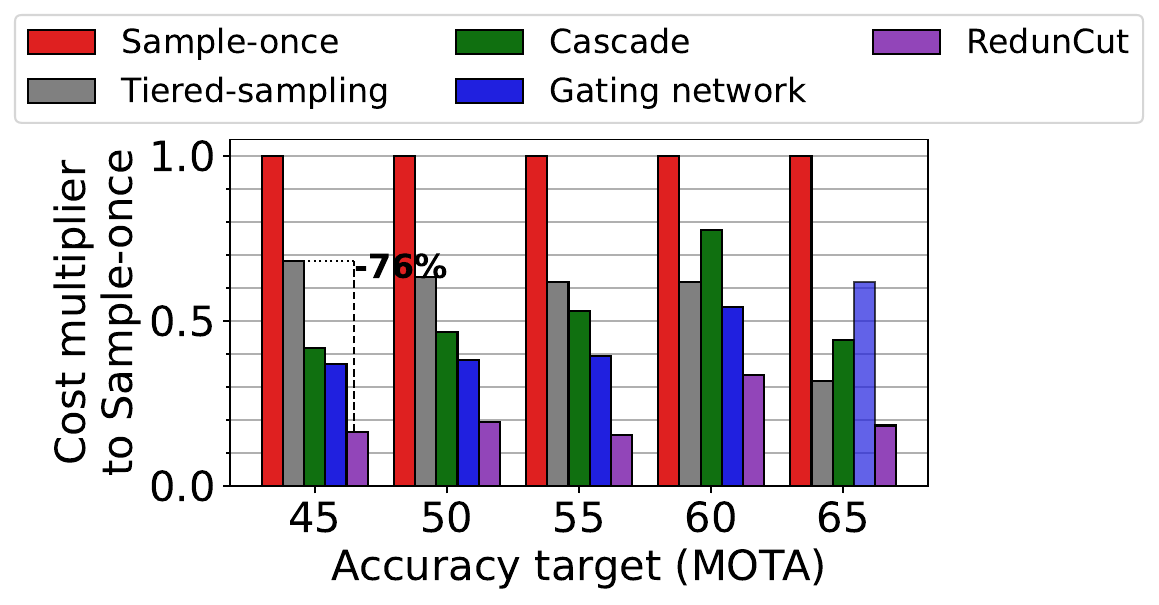}
    \caption{\small \textbf{The normalized compute cost of \system{} and other
            video analytics methods in surveillance camera videos.} Each method
            (color) is evaluated on a range of accuracy targets (x-axis). The
            resulting compute cost is normalized by dividing it by the compute
            cost of the sample-once method (y-axis). The single translucent
            blue bar is described in~\cref{ss:eval-main-results}.}
    \label{fig:barplot-meva-main-results}
\end{figure}

\begin{figure}[t!]
    \centering
    \includegraphics[width=0.45\textwidth,clip]{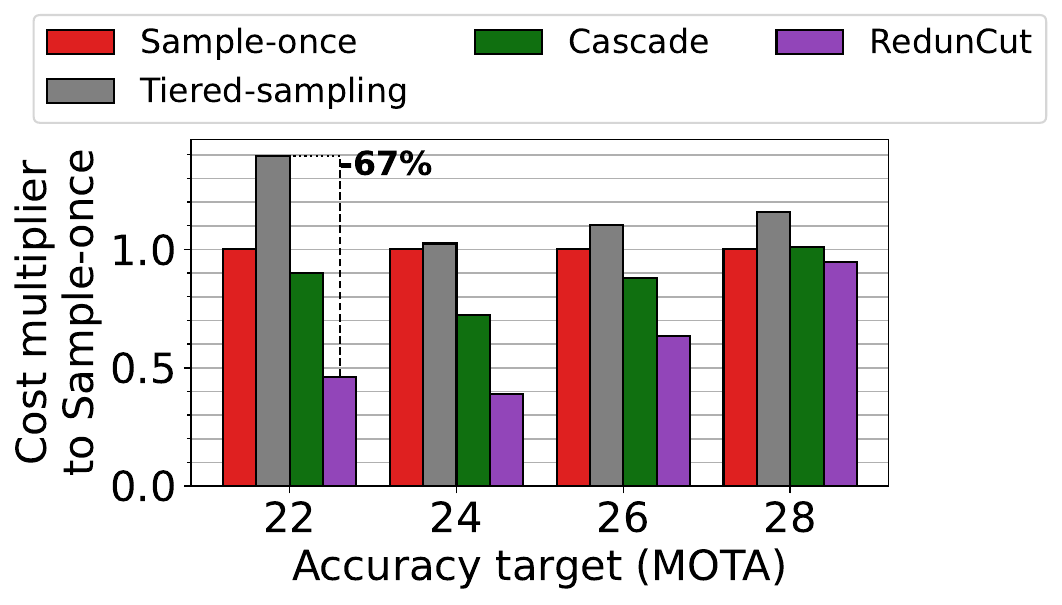}
    \caption{\small \textbf{The normalized compute cost of \system{} and other
            video analytics methods in drone camera videos.} Each method
            (color) is evaluated on a range of accuracy targets (x-axis). The
            resulting compute cost is normalized by dividing it by the compute
            cost of the sample-once method (y-axis).}
    \label{fig:barplot-visdrone-main-results}
\end{figure}

\Cref{fig:barplot-meva-main-results} and
\Cref{fig:barplot-visdrone-main-results} show the main results from
\Cref{fig:main-results} in barplot form.
\label{app:other-main-results}

\section{Spearman correlation between prediction statistics and accuracy}
\begin{figure}[t!]
    \centering
    \captionsetup[subfigure]{width=0.99\textwidth}
    \includegraphics[width=0.48\textwidth]{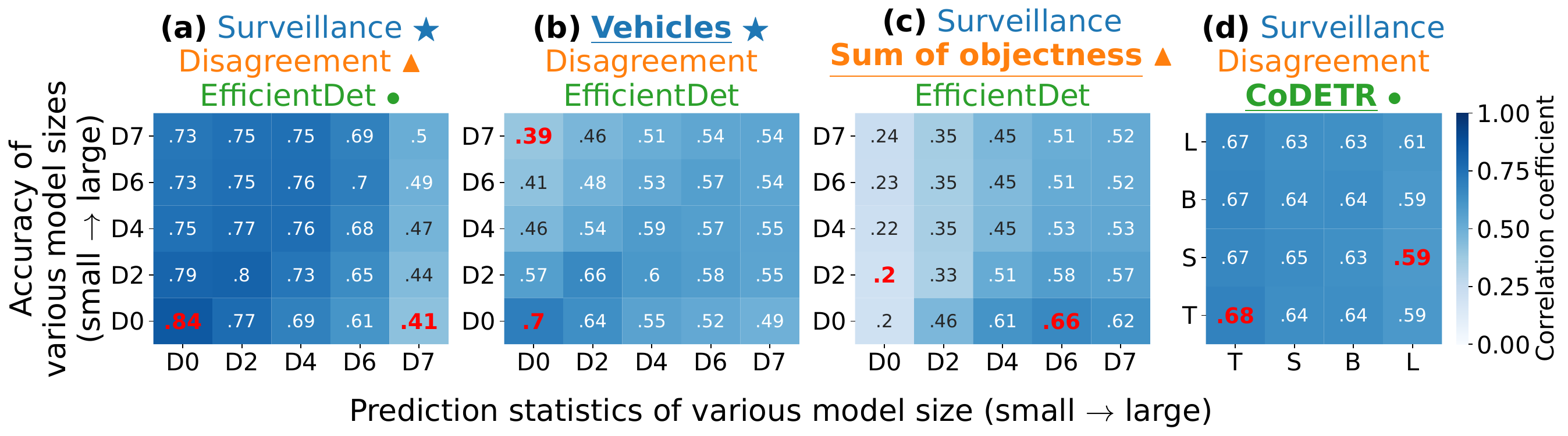}
    \caption{\small \textbf{Spearman correlation between the prediction statistics and the accuracy.}
    }
    \label{f:pred-stats-acc-corr-spearman}
\end{figure}

See \Cref{f:pred-stats-acc-corr-spearman}.
\label{app:pred-stats-acc-corr-spearman}

\section{Optimal number of models to sample in different settings}
\label{app:optimal-subset-sizes-other-settings}

\begin{figure}[t!]
    \centering
    \captionsetup[subfigure]{width=0.99\textwidth}
    \includegraphics[width=0.48\textwidth]{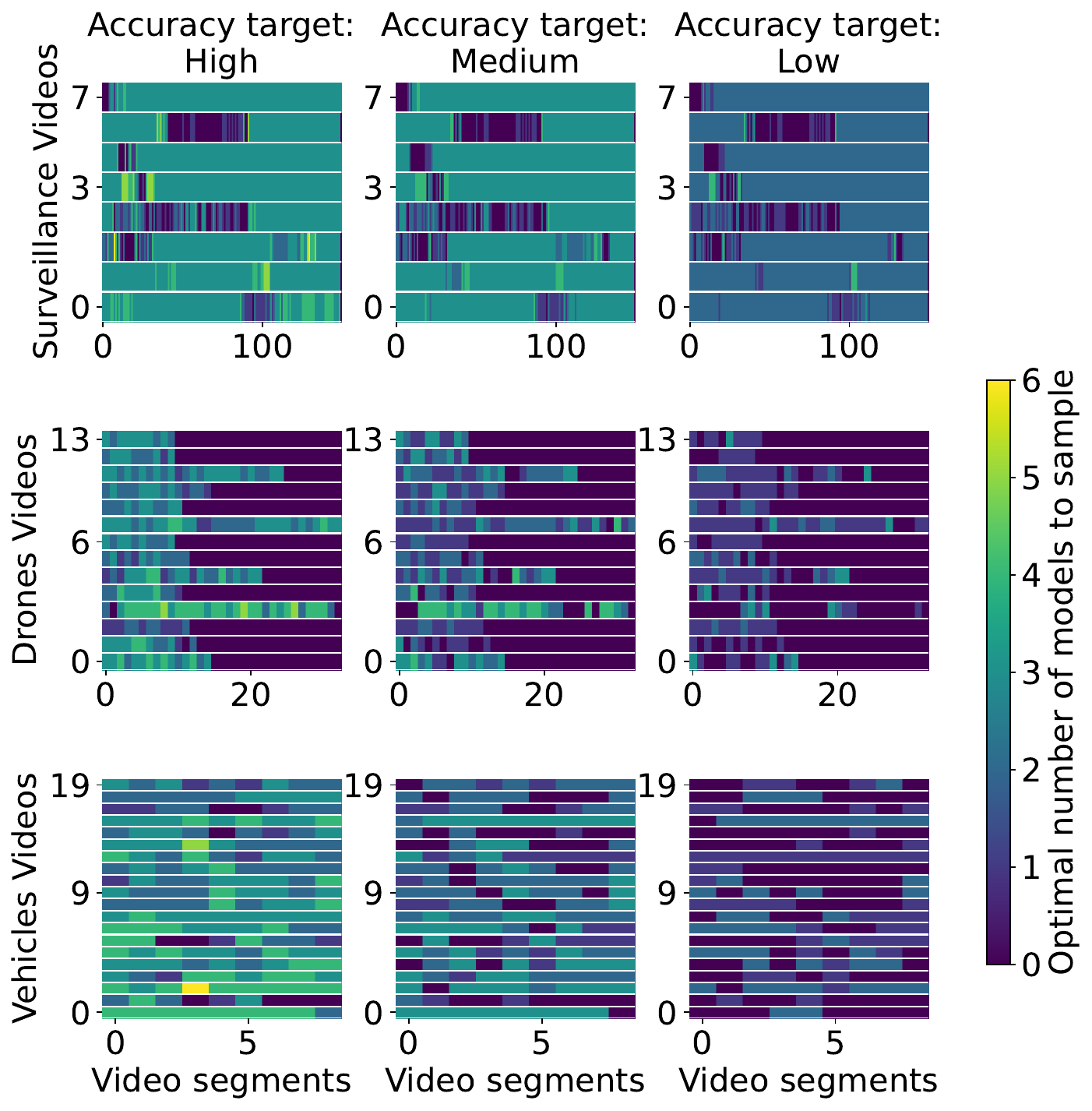}
    \caption{\small \textbf{The optimal number of models to sample in different accuracy targets and workload types.}
    }
    \label{f:optimal-sampling-subset-other-settings}
\end{figure}

See \Cref{f:optimal-sampling-subset} and
\Cref{f:optimal-sampling-subset-other-settings}.

\begin{figure}[t!]
    \centering
    \captionsetup[subfigure]{width=0.99\textwidth}
    \includegraphics[width=\linewidth]{figures/optimal_subset_sizes.pdf}
    \caption{\small \textbf{The optimal number of models to sample in
    different segments of videos.} The optimal number of models to sample
    (color) across different video segments (x-axis) of different videos
    (y-axis).}
    \label{f:optimal-sampling-subset}
\end{figure}

\section{Accuracy prediction error}
\label{app:accuracy-prediction-error}

\begin{figure}[t!]
    \centering
    \captionsetup[subfigure]{width=0.99\textwidth}
    \includegraphics[width=0.48\textwidth]{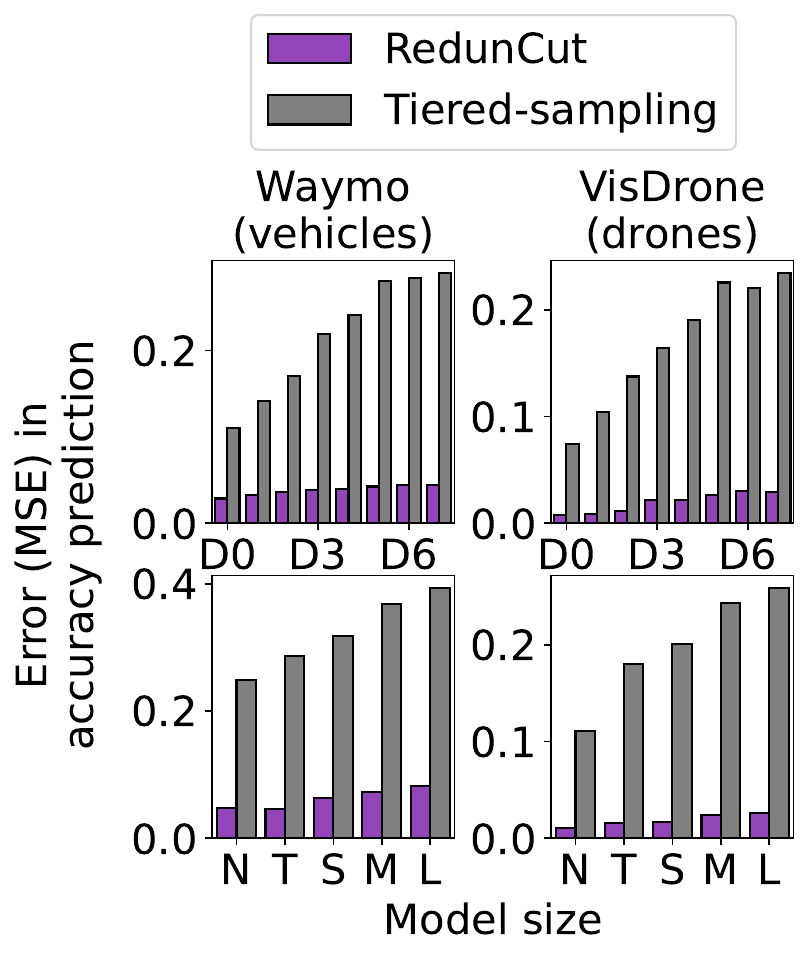}
    \caption{\small \textbf{The error of different accuracy prediction methods.}
    The error of accuracy prediction (Mean-squared error, y-axis) of different
    methods (color) across different datasets (left/right), model families
    (top/bottom), and different model sizes (x-axis). \system{}'s error is
    3.4-10$\times$ lower than the error that Tiered-sampling~\cite{chameleon}
    achieves using the model disagreement
    score~\cite{recl,ekya,videostorm,chameleon}.}
    \label{f:accuracy-prediction-error}
\end{figure}

See \Cref{f:accuracy-prediction-error}.

\section{Correlations between prediction statistics}
\label{app:pred-stats-pred-stats-corr}

\begin{figure}[t!]
    \centering
    \captionsetup[subfigure]{width=0.99\textwidth}
    \includegraphics[width=0.5\textwidth]{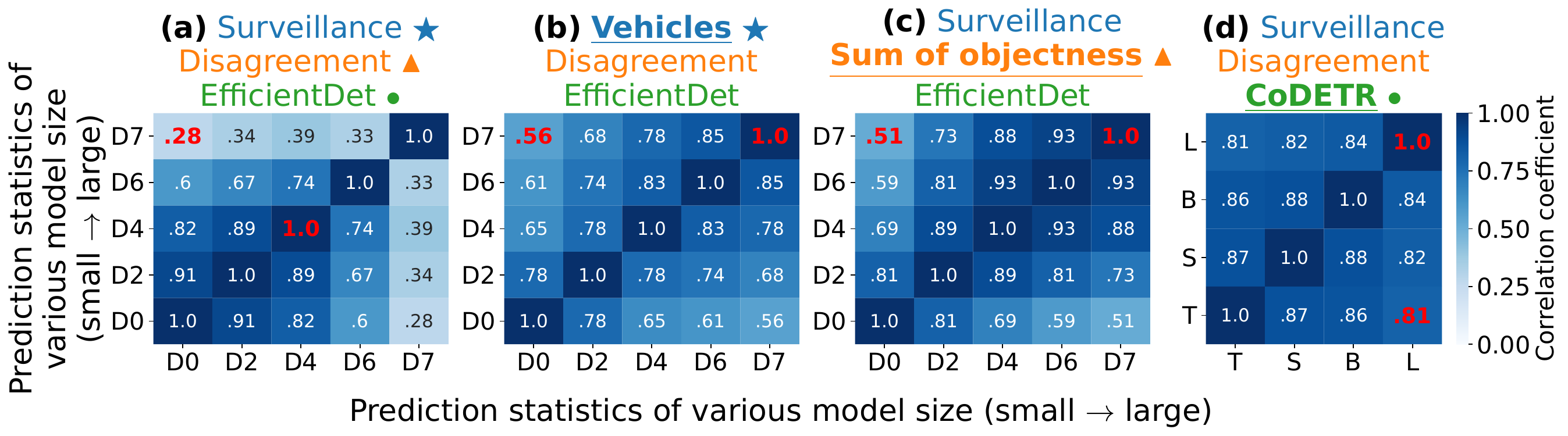}
    \caption{\small \textbf{Correlation among prediction statistics.} The
    correlation (Pearson) between the prediction statistics of different model
    sizes across workload types, prediction statistics types, and model
    families. See the caption of \Cref{f:pred-stats-acc-corr} for information
    about the headings and comparison between \textbf{(a)} - \textbf{(d)}.}
    \label{f:pred-stats-pred-stats-corr}
\end{figure}

See \Cref{f:pred-stats-pred-stats-corr}.

\section{Selector ablation}
\label{app:selector-ablation}

\begin{figure}[t!]
    \centering
    \captionsetup[subfigure]{width=0.99\textwidth}
    \includegraphics[width=0.5\textwidth]{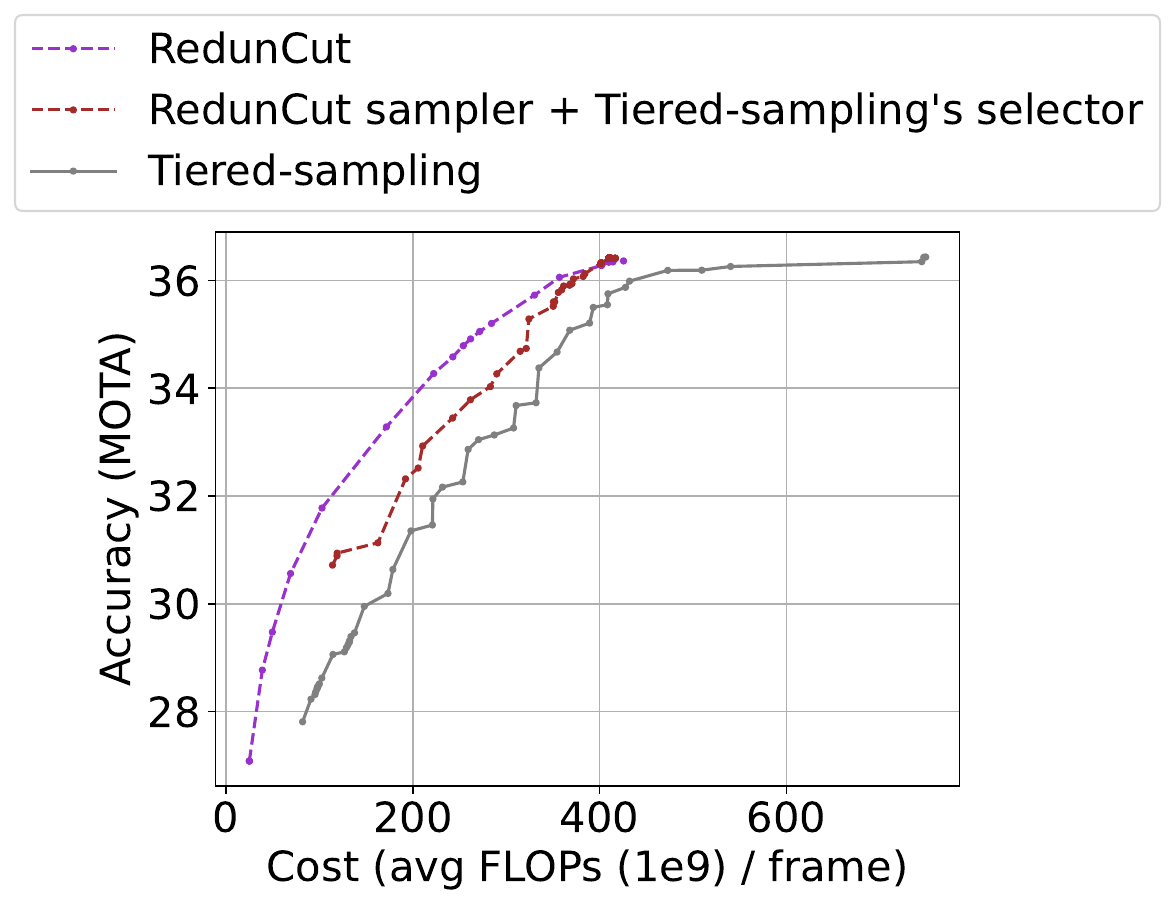}
    \caption{\small \textbf{An ablation of \system{}'s selection mechanism.}}
    \label{f:selector-ablation}
\end{figure}

See \Cref{f:selector-ablation}.

\end{appendix}

\end{document}